\documentclass[10pt,twocolumn,letterpaper]{article}

\usepackage[pagenumbers]{cvpr} 
\usepackage[misc,geometry]{ifsym}
\usepackage{graphicx}
\usepackage{amsmath}
\usepackage{amssymb}
\usepackage{booktabs}

\usepackage{blindtext}
\usepackage{multirow}
\usepackage{bm}
\usepackage{bbm} 
\usepackage{comment}
\usepackage{algorithm}
\usepackage{algorithmic}
\usepackage{array} 
\usepackage[dvipsnames]{xcolor}
\usepackage[safe]{tipa} 
\usepackage{color, colortbl}
\usepackage[misc,geometry]{ifsym} 

\makeatletter
\renewcommand\paragraph{
  \@startsection{paragraph} 
  {4} 
  {\z@} 
  {.5em \@plus1ex \@minus.2ex} 
  {-1.5em} 
  {\normalfont\normalsize\bfseries} 
}
\makeatother

\makeatletter
\def\@fnsymbol#1{\ensuremath{\ifcase#1\or \textsuperscript{~\Letter}\or \ddagger\or
   \mathsection\or \mathparagraph\or \|\or **\or \dagger\dagger
   \or \ddagger\ddagger \else\@ctrerr\fi}}
\makeatother

\definecolor{tabhighlight}{HTML}{e5e5e5}
\definecolor{citecolor}{HTML}{0071bc}

\usepackage[pagebackref,breaklinks,colorlinks,citecolor=citecolor]{hyperref}

\usepackage[capitalize]{cleveref}
\crefname{section}{Sec.}{Secs.}
\Crefname{section}{Section}{Sections}
\Crefname{table}{Table}{Tables}
\crefname{table}{Tab.}{Tabs.}

\begin{document}

\title{Chain of Thought Prompt Tuning for Vision-Language Models
}

\author{Jiaxin Ge\\
Peking University\\
 Beijing, China\\
{\tt\small aomaru@stu.pku.edu.cn}
\and
Hongyin Luo\\
MIT\\
Cambridge MA, US\\
{\tt\small hyluo@mit.edu}
\and
Siyuan Qian\\
Peking University\\
 Beijing, China\\
{\tt\small qiansiyuan@stu.pku.edu.cn}
\and
Yulu Gan\\
Peking University\\
 Beijing, China\\
{\tt\small ganyulu@stu.pku.edu.cn}
\and
Jie Fu\\
BAAI \\
Beijing, China\\
{\tt\small fujie@baai.ac.cn}
\and
Shanghang Zhang\thanks{Corresponding Author}\\
Peking University\\
 Beijing, China\\
{\tt\small shanghang@pku.edu.cn}
}

\maketitle

\maketitle

\footnote{This work is still ongoing and will be updated}

\begin{abstract}
   Language-Image Pre-training has demonstrated promising results on zero-shot and few-shot downstream tasks by prompting visual models with natural language prompts. However, most recent studies only use a single prompt for tuning, neglecting the inherent step-to-step cognitive reasoning process that humans conduct in complex task settings, for example, when processing images from unfamiliar domains.
   Chain of Thought is a simple and effective approximation to human reasoning process and has been proven useful for natural language processing (NLP) tasks.
   Based on this cognitive intuition, we believe that conducting effective reasoning is also an important problem in visual tasks, and a chain of thought could be a solution to this problem. In this work, we propose a novel chain of thought prompt tuning for vision-language modeling.
   Extensive experiments show that our method not only generalizes better in image classification tasks, has greater transferability beyond a single dataset, and has stronger domain generalization performance, but also performs much better in image-text retrieval and visual question answering, which require more reasoning capabilities.
   We are the first to successfully adapt chain-of-thought prompting that combines visual and textual embeddings.  
   We will release our codes. 
\end{abstract}

\section{Introduction}
\label{sec:intro}
\begin{figure}[t]
   \center{\includegraphics[width=8cm]  {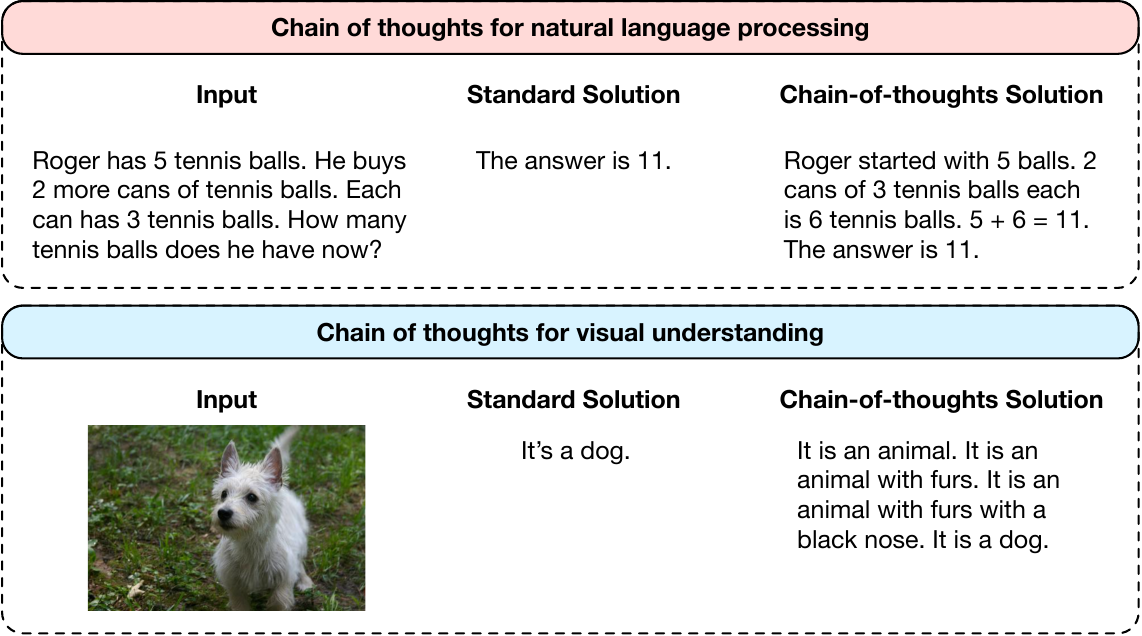}} 
   \caption{A conceptual overview of the chain of thought prompt tuning. We compare chain-of-thought prompt construction for language and vision understanding tasks.}
   \label{demonstration}
\end{figure}
Multi-modal methods with pretrained vision and language models, including CLIP\cite{CLIP}, ALIGN \cite{Lietal2021}, and Flamingo \cite{Flamingo}, have achieved significant results in vision-text matching.
The vision-language models contain an image encoder and a text encoder, which are trained contrastively and can effectively align semantically related concepts in a shared embedding space.
With the ability to produce classification weights diametrically, the text encoder can effectively guide the model to predict accurately in zero-shot scenarios, including tasks like image classification \cite{Imagenetfeifei}, visual question answering \cite{VQA}, and image-text retrieval\cite{Retrieval}. 
However, manually creating prompts for vision and language models is a laborious procedure that often requires previous expert knowledge. 
Furthermore, if the manually designed prompts are not appropriate for the task, the model's performance will be impaired.

To automate the prompt engineering process, \cite{zhouetal2022cocoop,Lesterprompt,mscoco} propose to optimize prompts using data from downstream tasks by minimizing a cross entropy loss.
Contrary to manual design, this method frees us from human prompt engineering and can effectively learn prompts that adapt well to downstream tasks, avoiding the issue that human-designed prompts might not be optimal for the task. 
Although achieving a good result on the specific downstream tasks, prompt tuning has a detrimental effect on the vision-language models' capacity to generalize, making them unable to adapt well to unseen visual concepts. 
Bridging this generalization gap is an urgent issue and attracts much attention in the community\cite{zhouetal2022cocoop,Zhupromptaligned}. 

Unfortunately, these works use natural language as guidance to solve vision tasks, but they all fail to notice a fundamental yet very important feature in natural language: the power of reasoning. 
Cognitively, the human brain is the only resilient system that can adapt successfully to new circumstances quickly, and many studies in different domains have shown the importance of studying how human brains operate and mimicking their behaviors to improve the generalizing capabilities of a model\cite{gwt,inductivebias,dualreasoning}.
Recently, it has been shown in the NLP community that training a model with additional guidance on human-like reasoning processes can considerably improve the generalizing capabilities of a model \cite{JasonWei2022ChainOT}. 
Rather than directly generating the final solution, generating the answer with a chain of reasoning sentences could probe the reasoning capacity of language models and hence generalize better for addressing different problems. A demonstration is shown in Figure \ref{demonstration}, when answering a natural language question, instead of directly generating "the answer is 11", the language model is trained to generate a chain of reasoning sentences like "Roger started with 5 balls", "2 cans of tennis balls each is 6 tennis balls.", "5+6=11.", "The answer is 11." Similarly, when given an image input, instead of asking the model to generate "It is a dog", we want it to go through a reasoning process like "It is an animal.", "It is an animal with fur.", "It is an animal with furs and a black nose.", "It is a dog."

Previous studies on vision-language models rely on a single prompt to guide vision tasks, which could not replicate the intrinsic reasoning process that people use, especially in new situations. In this work, we introduce the chain of thought intuition for visual understanding tasks. Although this intuition is straightforward in NLP since one can express their thought process directly with a few sentences and reasoning with language is natural, it is difficult to use in vision, which requires effectively combining step-by-step visual information extracted from an image with language reasoning.

To overcome this difficulty, we propose a novel architecture to emulate this chain-of-thought reasoning process and apply it to vision tasks. 
First, we design multiple prompts that are connected together to form a chain, with each prompt receiving information from the previous prompt and passing it on to the next prompt. Then, instead of using a single network to generate an input-specific bias for the prompt, we design a chain of networks, where each network generates a step-specific bias corresponding to a specific step in a reasoning chain. Finally, we add a dynamic chain controller that is used to control the chain dynamically based on the input.



We perform extensive experiments on a variety of tasks, including image classification, image-text retrieval, and visual question answering. In image classification, we conduct tasks on base to new generalization using 11 datasets covering a wide range of subjects. We also conduct a cross-dataset transfer experiment, in which the model is trained on ImageNet and tested directly on the other 10 datasets. In addition, We carry out a domain generalization experiment, in which the model trained on ImageNet is tested directly on other ImageNet datasets. In image-text retrieval, we conduct experiments on MSCOCO\cite{mscoco} and Flickr30k\cite{Retrieval}. Finally, in visual question answering, we test on the VQAv2\cite{VQA} dataset. On all tasks, we attain a significant performance gain, which demonstrates the superiority of our approach. We also test the impact of our chaining architecture, and the experimental results prove that the chaining structure of either the prompts or the Meta-Nets is beneficial for out-of-domain generalization.

To the best of our knowledge, this paper makes the following novel contributions:
\begin{itemize}\setlength{\itemsep}{0pt} \setlength{\parsep}{0pt}
    \item We are the first to successfully adapt chain-of-thought for prompt tuning that combines visual and textual embeddings in vision domain. It is consistent with the human learning paradigm, providing unique insight in vision domain.
    
    \item We propose a novel architecture to mimic the chain of thought processes of human. It effectively models the relation between prompts and is simple yet effective, requiring minimal additional computing resources and is easy to plug in, indicating its great potential for various applications.
    
    \item Demonstrates consistent performance inprovement by applying Chain of Thought architecture on five tasks. For base to new generalization, we improve CoCoOp by over 1.27\%. For cross dataset transfer and domain generalization, we improve by 0.5\% and 0.26\%. For image-text retrieval, we improve by 0.97\%, and finally for VQA, we improve by 1.07\%.
\end{itemize}

\begin{figure*}[t]
   \center{\includegraphics[width=15cm]  {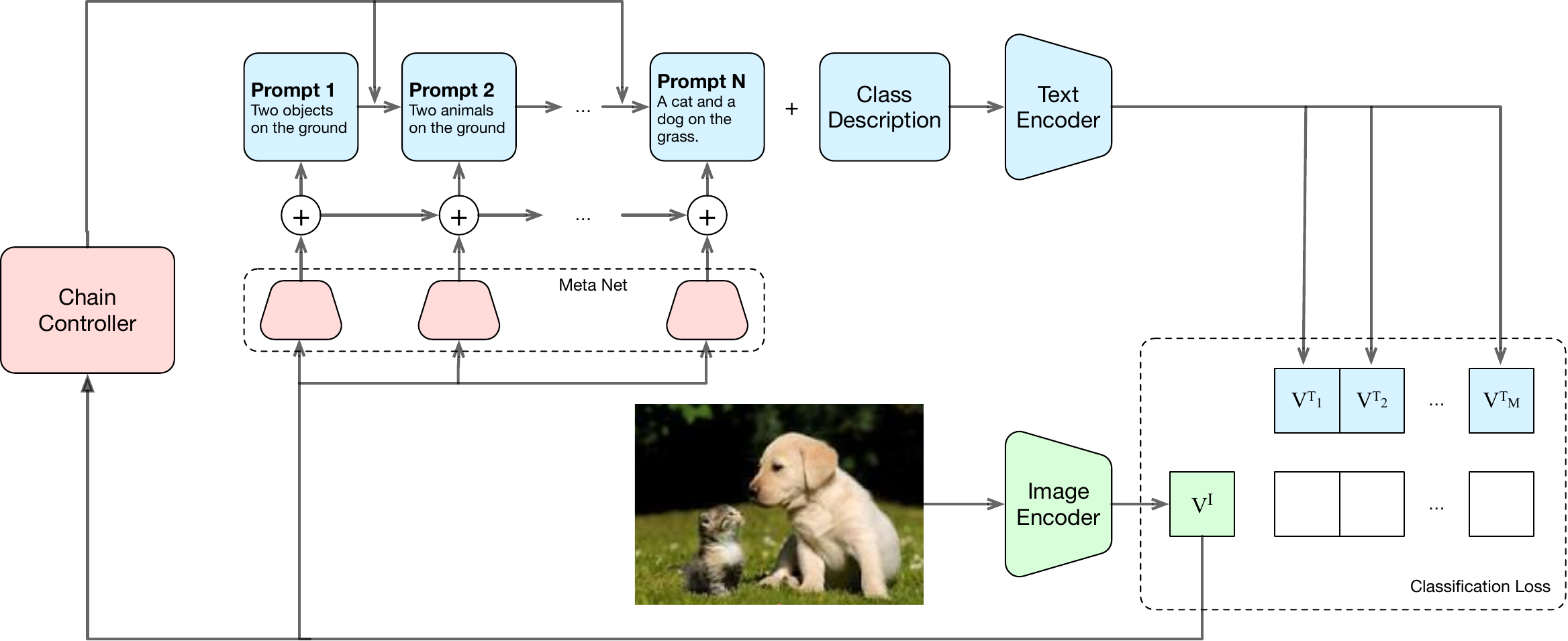}} 
   \caption{The multi-modal chain of thought prompt tuning framework. We build a series of chained prompts and a sequence of Meta-Nets. We also adopt a dynamic chain controller to control the weights on the chain based on the inputs. We use the final prompt (prompt N) for prediction.}
   \label{framework}
\end{figure*}
\section{Related Works}
\subsection{Vision Language Models}
Using natural language supervision for computer vision has drawn a wide range of attention. 
Inspired by cross-domain techniques, recent studies\cite{CLIP, zhouetal2022cocoop, zhouetal2022coop} align images and texts to learn a joint embedding space. 
Specifically, V-L models like CLIP \cite{CLIP}, LiT \cite{XiaohuaZhai2021LiTZT}, FILIP \cite{LeweiYao2022FILIPFI},  ALIGN \cite{Lietal2021}, and Florence Yuan et al. (2021) \cite{florence} demonstrate exceptional performance on a variety of tasks,  especially in few-shot or zero-shot settings. These models are pre-trained using a large number of text-image pairs online, bridging the gaps between text and image modalities.

Although these pre-trained models are intriguing, efficiently adapting them to downstream tasks remains crucial yet difficult. Many studies have been conducted to learn an effective prompt for downstream tasks\cite{XinZhang2021DomainPL,YuningLu2022PromptDL}, but none of them have caught the implicit reasoning logic in prompts. We enhance the performance of prompts through a simple yet effective chaining network.
\subsection{Prompt Tuning in Vision-Language Models}
prompt tuning is first proposed in natural language processing (NLP). 
It treats the downstream datasets as a language mask problem so that the frozen language model can be directly adapted to new tasks \cite{Gaoetal2021}, keeping the embedding space that it has learned from pre-training. 
Recent works including \cite{ZifengWang2022LearningTP} generate a group of candidate prompts, produce a set of candidate prompts from which the best ones are picked for the highest training accuracy. AutoPrompt \cite{TaylorShin2020AutoPromptEK} elicits knowledge from language models with automatically generated discrete prompts. Other works like \cite{ShoufaChen2022AdaptFormerAV} use prompt tuning to efficiently adapt pre-trained ViTs to a variety of image and video tasks.
Prompt tuning also receives great interest in the computer vision domain.
CoOp \cite{zhouetal2022coop} introduces prompt tuning for vision tasks. 
They adapt it to the pre-trained visual-language model by transforming the context word into a set of learnable vectors. CoCoOp \cite{zhouetal2022cocoop} further changes this static prompt to a dynamic prompt, allowing it to better handle class shifts. 

The key difference between our work and the above work, especially for prompt tuning in vision-language tasks, is that we simulate the human step-by-step thinking process by considering multiple prompt steps.
\subsection{Chain of Thought}
The concept of chain-of-thought is first introduced in the NLP domain. \cite{JasonWei2022ChainOT} first proposes chain-of-thought prompting, showing that adding a series of intermediate reasoning steps could significantly improve the ability of large language models to perform reasoning. Many subsequent NLP works are based on this concept. \cite{innermonologue} proposes to use the reasoning process by language models to plan and interact with robots. \cite{humanreason} further develops machine reasoning by examining a more human-like thinking process. 

Our work draws on the idea of the chain of thought. In the visual-language task, we consider each step to have a different degree of importance when reasoning according to the steps of thought. In addition, the previous steps are easily forgotten afterward. Our proposed approach takes the above two factors into account to make it more reasonable to perform visual language tasks.

Closest to our work is \cite{learntoexplain}, which applies chain of thought in visual question answering. However, their approach only uses a context explanation of an image that is pre-obtained and then feeds the context into a GPT-3\cite{GPT3} language model. Their main contribution is still limited to adopting chain of thought to language models, ignoring the visual features extracted from the image itself. Our work, however, is able to combine visual and textual embeddings.

\section{Method}
In this section, we first formulate the problem and then provide a brief overview of our method. then, we sequentially introduce the three main components in our architecture: Prompt Chaining, Self Adaptive Chain Controller, and Meta-Nets Chaining.
\subsection{Problem Formulation}
Our goal is to learn a $chain$ of generalizable prompt representations via limited data.
Existing solutions (e.g., CLIP\cite{CLIP}, CoCoOp\cite{zhouetal2022cocoop}, CoOp\cite{zhouetal2022coop}) first extract visual feature $v$ of each input image by feeding it into CLIP’s vision encoder, and next extract text embeddings $t$ of each class by feeding the according prompt into the CLIP's text encoder. Then, the probability of each class is computed as
\begin{equation}
  p(t_i|x) = \frac {e^{<G(t_i), v> / \tau}}{\sum_{i} e^{<G(t_i), v>/ \tau}}
\end{equation}
where $x$ stands for the input image, $\tau$ is the temperature parameter and $< , >$ stands for the cosine similarity between two embeddings. $v$ is the visual embedding of the input image, and $G(t_i)$ stands for the text embedding of concatenated prompt and class description sentences. Then, applying the cross-entropy loss, gradients are back-propagated to update the prompt. During training, the weights of the text encoder and the vision encoder are frozen to preserve the representation space that they learned from pre-training.
\subsection{Method Overview}
An Overview of our approach is shown in Figure \ref{framework}. 
Our primary motivation is to decompose the process of understanding the content of a picture into a step-to-step reasoning process.
To this end, we design a chain of prompts. Each prompt receives information from its prior prompt, and then passes its information to the next prompt.

Because different images require quite different reasoning processes, we believe that adding control to the chain based on the input is necessary. Therefore, we design a control module to dynamically control the chain based on image inputs. 
Finally, we use the prompt N, the last prompt in the chain, for prediction, as the final step of a reasoning process is to draw a conclusion.

As CoCoOp has demonstrated, adding a light-weighted neural network to dynamically encode the visual feature and add it to the prompt could greatly enhance the network's performance in OOD settings. Instead of using a single network for encoding, we hypothesize that using a chain of networks could be beneficial, because each network could learn to encode the visual feature in a step-specific way. To preserve the initial information and avoid vanishing gradients, we feed the original visual feature into every network. 
We design the chain in a residual way, where the output of the prior network is added directly to the output of the current network.

Each network corresponds to a specific prompt, allowing each prompt to acquire a step-specific bias.

\subsection{Chain of Thought: Prompt Chaining}
Our main motivation is to decompose an image recognition process into a step-by-step cognitive reasoning process. To mimic the thinking process, we build a set of prompts that are chained together, each prompt receiving information from the prior one and sending information to the next one.
Assume that we have a total of $C$ classes, and the label of class $i$ is $h_i$. Then, the prompt for class $i$ at the $j$-step in the chain is a concatenated sentence $t_j^i= (p_j, h_i)$, where $p_j$ is the task-irrelevant prompt generated at the $j$-th step in the prompt chain.
The reasoning process uses an averaged embedding of the prompt flow of thought. Therefore, during the forward pass, each prompt embedding is a combination of its own and its prior:
\begin{equation}
  G(t_j^i) = (1-\lambda_{j})*G(t_{j-1}^i) + \lambda_{j}*G(t_j^i)
\end{equation}
where $\lambda_{j}$ is a step-specific parameter generated by the chain controller. Using this method, each prompt can acquire the outcome of the reasoning from its previous step and conduct its own reasoning in the current step. 
In this way, our chained prompts preserve an inherent reasoning ability. When encountering new scenarios, this reasoning ability aids the model to deduct and infer new concepts that it hasn't seen before. Thus, the model is more robust when seeing new visual concepts or when solving new tasks.
\subsection{Self Adaptive Chain Controller}
Intuitively, different tasks require different reasoning processes. Easier image classification tasks might need a shorter and less complicated chain, while harder tasks like image-text retrieval might need more complicated chains. Moreover, some images are straightforward and easier to recognize, while other photos might need more reasoning to recognize.
Therefore, we add a self-adaptive chain controller module to dynamically control the chain on the basis of the input. 
More specifically, for each image instance, the visual features of the image are fed into the chain controller. Then, the chain controller outputs the parameters on the chain, from $\lambda_{1}$ to $\lambda_{n}$.
Since we want to introduce a method that could be easily plugged in in any setting, we use a light-weighted chain controller, with a simple structure: linear-relu-linear-sigmoid. We empirically find that by introducing such a small network with just a few additional parameters, the chain could adapt well to many new scenarios while preserving its stability.
\subsection{Meta-Nets Chaining}
First, we briefly review the Meta-Net proposed in CoCoOp\cite{zhouetal2022cocoop}. 
CoCoOp adds a light weighted neural network(namely, Meta-Net) to CoOp\cite{zhouetal2022coop} that could encode the visual feature v produced by the image encoder. The output is added to the prompt as a bias to help the prompt adapt to each instance dynamically. CoCoOp has demonstrated that this alternation could greatly promote the prompt's generalizing ability.

However, CoCoOp's single network fails to model a step-by-step reasoning process that human conducts. To overcome this deficiency, we design a chain of networks.

During the reasoning process, we utilize a different subset of visual information at each step with a meta-net. Therefore, we calculate a visual representation as a bias term that is added to the input word embeddings at each step. To this end, we build a chain of networks to generate the visual bias, where each network corresponds to a specific step. Specifically, the output of the $j$-th meta net is the bias of step $j$ and is added to the word embeddings of prompt $j$ before being fed into the text encoder. 
\begin{equation}
  \hat{E}_j = E_j + v_j
\end{equation}
Where $\hat{E}_j$ is the updated word embeddings of the text encoder, $\hat{E}_j$ is the original word embeddings, and $v_j$ stands for the visual embedding. The networks should also form a chain relation yet not lose the original information, so we let each network receive information from the previous network and from the original image. We find out that to better preserve information, avoid gradient vanishing, and enhance stability, a residual-like architecture is preferred. The output from the previous network is added directly to the output from the currently network, promoting a stabler architecture.

\section{Experiments}
\begin{table*}
 \centering
 \caption{Base To New Results. We compute base-class accuracy, novel-class accuracy, and the Harmonic mean of the base-class and the novel-class. The Harmonic mean(H) is a metric also compared in CoCoOp that could measure the generalization trade-off. } \label{table_1}
 \footnotesize
 \begin{minipage}{0.32\textwidth}
  \centering (a) Average \\
  \begin{tabular}{ccc|c}
   \toprule
   & Base & New & H \\
   \midrule
   CLIP &  69.34 & \textbf{74.22} & 71.70  \\
   CoOp & \textbf{82.69} & 63.22 & 71.66 \\
   CoCoOp & 80.47 & 71.69 & 75.83 \\
   \cellcolor{lightgray}Ours & \cellcolor{lightgray}80.23 & \cellcolor{lightgray}74.20 & \cellcolor{lightgray}\textbf{77.10} \\ \bottomrule
  \end{tabular}
 \end{minipage}
 \begin{minipage}{0.32\textwidth}
  \centering  (b) ImageNet\\
  \begin{tabular}{ccc|c}
   \toprule
   & Base & New & H \\ \midrule
   CLIP & 72.43 & 68.14 & 70.22 \\
   CoOp & \textbf{76.47} & 67.88 & 71.92 \\
   CoCoOp & 75.98 & 70.43 & 73.10 \\
   \cellcolor{lightgray}Ours & \cellcolor{lightgray}76.00 & \cellcolor{lightgray}\textbf{70.68} & \cellcolor{lightgray}\textbf{73.24}\\\bottomrule
  \end{tabular}  
 \end{minipage}
 \vspace{0.1in}
 \begin{minipage}{0.32\textwidth}
  \centering (c) Caltech101\\
  \begin{tabular}{ccc|c}
   \toprule
   & Base & New & H \\
   \midrule
   CLIP &  96.84 & 94.00 & 95.40  \\
   CoOp & \textbf{98.00} & 89.81 & 93.73 \\
   CoCoOp & 97.96 & 93.81 & 95.84 \\
   \cellcolor{lightgray}Ours & \cellcolor{lightgray}97.91 & \cellcolor{lightgray}\textbf{94.03} &\cellcolor{lightgray}\textbf{95.93} \\\bottomrule
  \end{tabular}
 \end{minipage}
 \vspace{0.1in}
 \begin{minipage}{0.32\textwidth}
  \centering (d) OxfordPets\\
  \begin{tabular}{ccc|c}
   \toprule
   & Base & New & H \\
   \midrule
   CLIP & 91.17 & 97.26 & 94.12 \\
   CoOp &  93.67 & 95.29 & 94.47 \\
   CoCoOp & 95.20 & 97.69 & 96.43 \\
   \cellcolor{lightgray}Ours & \cellcolor{lightgray}\textbf{95.43} & \cellcolor{lightgray}\textbf{97.78} & \cellcolor{lightgray}\textbf{96.59} \\\bottomrule
  \end{tabular}
 \end{minipage}
 \begin{minipage}{0.32\textwidth}
  \centering (e) StanfordCars\\
  \begin{tabular}{ccc|c}
   \toprule
   & Base & New & H \\
   \midrule
   CLIP &  63.37 & \textbf{74.89} & 68.65 \\
   CoOp & \textbf{78.12} & 60.40 & 68.13 \\
   CoCoOp & 70.49 & 73.59 & 72.01 \\
   \cellcolor{lightgray}Ours & \cellcolor{lightgray}70.59 & \cellcolor{lightgray}73.82 &  \cellcolor{lightgray}\textbf{72.17}\\\bottomrule
  \end{tabular}
 \end{minipage}
 \begin{minipage}{0.32\textwidth}
  \centering (f) Flowers102\\
  \begin{tabular}{ccc|c}
   \toprule
   & Base & New & H \\
   \midrule
   CLIP &  72.08 & \textbf{77.80} & 74.83 \\
   CoOp & \textbf{97.60} & 59.67 & 74.06\\
   CoCoOp & 94.87 & 71.75 & 81.71\\
   \cellcolor{lightgray}Ours & \cellcolor{lightgray}94.46 & \cellcolor{lightgray}72.46 & \cellcolor{lightgray}\textbf{82.01}  \\ \bottomrule
  \end{tabular}
 \end{minipage}
 \vspace{0.1in}
 \begin{minipage}{0.32\textwidth}
  \centering (g) Food101\\
  \begin{tabular}{ccc|c}
   \toprule
   & Base & New & H \\
   \midrule
   CLIP & 90.10 & 91.22 & 90.66 \\
   CoOp &  88.33 & 82.26 & 85.19 \\
   CoCoOp & 90.70 & 91.29 & 90.99 \\
   \cellcolor{lightgray}Ours & \cellcolor{lightgray}\textbf{90.74} & \cellcolor{lightgray}\textbf{91.77} & \cellcolor{lightgray}\textbf{91.25}\\\bottomrule
  \end{tabular}
 \end{minipage}
 \begin{minipage}{0.32\textwidth}
  \centering (h) FGVCAircraft
  \begin{tabular}{ccc|c}
   \toprule
   & Base & New & H \\
   \midrule
   CLIP & 27.19 & \textbf{36.29} & 31.09 \\
   CoOp & \textbf{40.44} & 22.30 & 28.75 \\
   CoCoOp &  33.41 & 23.71 & 27.74 \\
   \cellcolor{lightgray}Ours & \cellcolor{lightgray}35.13 & \cellcolor{lightgray}32.21 & \cellcolor{lightgray}\textbf{33.61}\\\bottomrule
  \end{tabular}
 \end{minipage}
 \begin{minipage}{0.32\textwidth}
  \centering (i) SUN397 \\
  \begin{tabular}{ccc|c}
   \toprule
   & Base & New & H \\
   \midrule
   CLIP &  69.36 & 75.35 & 72.23 \\
   CoOp & \textbf{80.60} & 65.89 & 72.51 \\
   CoCoOp & 79.74 & 76.86 & 78.27 \\
   \cellcolor{lightgray}Ours & \cellcolor{lightgray}79.44 & \cellcolor{lightgray}\textbf{77.20} & \cellcolor{lightgray}\textbf{78.30} \\ \bottomrule
  \end{tabular}
 \end{minipage}
 \vspace{0.1in}
 \begin{minipage}{0.32\textwidth}
  \centering (j) DTD \\
  \begin{tabular}{ccc|c}
   \toprule
   & Base & New & H \\
   \midrule
   CLIP &  53.24 & \textbf{59.90} & 56.37 \\
   CoOp & \textbf{79.44} & 41.18 & 54.24 \\
   CoCoOp & 77.01 & 56.00 & 64.85\\
   \cellcolor{lightgray}Ours & \cellcolor{lightgray}76.27 & \cellcolor{lightgray}58.34 & \cellcolor{lightgray}\textbf{66.11} \\\bottomrule
  \end{tabular}
 \end{minipage}
 \begin{minipage}{0.32\textwidth}
  \centering (k) EuroSAT \\
  \begin{tabular}{ccc|c}
   \toprule
   & Base & New & H \\
   \midrule
   CLIP & 56.48 & 64.05 & 60.03 \\
   CoOp &  \textbf{92.19} & 54.74 & 68.69 \\
   CoCoOp &  87.49 & 60.04 & 71.21 \\
   \cellcolor{lightgray}Ours & \cellcolor{lightgray}84.11 & \cellcolor{lightgray}\textbf{72.81} & \cellcolor{lightgray}\textbf{78.06}  \\\bottomrule
  \end{tabular}
 \end{minipage}
 \begin{minipage}{0.32\textwidth}
  \centering (l) UCF101\\
  \begin{tabular}{ccc|c}
   \toprule
   & Base & New & H \\
   \midrule
   CLIP &  70.53 & \textbf{77.50} & 73.85 \\
   CoOp & \textbf{84.69} & 56.05 & 67.46 \\
   CoCoOp &  82.33 & 73.45 & 77.64 \\
   \cellcolor{lightgray}Ours & \cellcolor{lightgray}82.47 & \cellcolor{lightgray}75.09 & \cellcolor{lightgray}\textbf{78.61}\\ \bottomrule
  \end{tabular}
 \end{minipage}
 \label{tab:temps}
\end{table*}
We conduct extensive experiments on five primary tasks using a total of 18 datasets and four ablation experiments that validate the effectiveness of our architecture. We provide a detailed description of benchmark settings and quantitive results analysis.
\subsection{Tasks}
\textbf{Base-to-New Generalization}
The first experiment we conduct on image classification is a generalization from Base to New classes. We follow a zero-shot setting where the classes of a dataset are equally split into base classes and novel classes. The model is only trained on the base classes and then tested on both the base and new classes. We report the base-class accuracy, the novel-class accuracy, and their harmonic mean (H). Consistent with CoCoOp\cite{zhouetal2022cocoop}, we especially highlight the H score, which measures a generalization trade-off and is thus a more comprehensive measurement. 

\textbf{Cross-dataset Evaluation}
The second experiment we conduct on image classification is cross-dataset evaluation. In this setting, the model is trained on the ImageNet dataset\cite{Imagenetfeifei} and then directly evaluated on ten other datasets. Consistent with CoCoOp, our model is trained on the full Imagenet, which embodies 1000 classes. This task is a more challenging domain generalization problem.

\textbf{Domain Generalization}
The final experiment we conduct on image classification is domain generalization.
Following the benchmarks in \cite{zhouetal2022cocoop, zhouetal2022coop}, we use the model trained on the full ImageNet and test it directly on other four ImageNet datasets that contains various domain shifts. 

\textbf{Image-Text Retrieval}
Aside from the image classification tasks that are most widely used, we also test our approach on more callenging visual tasks that require more complicated reasoning. In this experiment, we test our method on two image-text retrieval captioning datasets, Flickr30k\cite{Retrieval} and MSCOCO\cite{mscoco}. We sample a subset from the dataset using Karpathy split and train the model on the subset. Then, we test the model on the test set. We use the metric Recall at 1 (R@1).

\textbf{Visual Question Answering}
Lastly, we use the VQAv2\cite{VQA} dataset to adopt the visual question answering task. Following \cite{vqa-caption}, we consider visual question answering as classification problem where the model chooses the answer from a set of delegate answers given a question. This task requires a great amount of reasoning, which is compatible with our intuition.
\begin{table*}
\centering
 \footnotesize
    \caption{Cross Dataset Transfer Result. We train the model on the full ImageNet and then test its performance directly on other ten datasets. Our model outperforms CoOp and CoOp on eight out of ten datasets. Demonstrating a strong generalizating ability.}
    \begin{tabular}{cccccccccccc}
    \toprule
     &
    \rotatebox{90}{Caltech101}&
    \rotatebox{90}{OxfordPets}&
    \rotatebox{90}{StanfordCars}&
    \rotatebox{90}{Flowers102}&
    \rotatebox{90}{Food101}&
    \rotatebox{90}{FGVCAircraft}&
    \rotatebox{90}{SUN397}&
    \rotatebox{90}{DTD}&
    \rotatebox{90}{EuroSAT}&
    \rotatebox{90}{UCF101}&
    \rotatebox{90}{Average} \\
     \midrule
    CoOp & 93.70 & 89.14 & 64.51 & 68.71 & 85.30 & 18.47 & 64.15 & 41.92 & 46.39 & 66.55 & 63.88 \\
    CoCoOp & 94.43 & 90.14 & 65.32 & 71.88 & 86.06 & 22.94 & \textbf{67.36} & \textbf{45.73} & 45.37 & 68.21 & 65.74 \\
    Ours  & \textbf{94.48} & \textbf{90.79} & \textbf{65.35} & \textbf{72.77} & \textbf{86.37} & \textbf{23.52} & 67.29 & 44.44 & \textbf{48.86} & \textbf{68.49} & \textbf{66.24} \\ \bottomrule
    \end{tabular}
    \label{cross_dataset}
\end{table*}

\begin{table*}
\centering
\caption{Domain Generalization Experiment. The model is trained on the full ImageNet and then tested directly on four other ImageNets that contains various domain shifts, including ImageNetv2, ImageNet-Sketch, ImageNet-a, and also ImageNet-r. }
\footnotesize
\setlength\tabcolsep{10pt}
    \begin{tabular}{cccccc}
    \toprule
     &
    ImageNetV2 & ImageNet-Sketch & ImageNet-A & ImageNet-R
    & Average  \\
     \midrule
    CLIP &   60.83 & 46.15 & 47.77 & 73.96 & 57.18  \\
    CoOp & 64.20 & 47.99 & 49.71 & 75.21 & 59.28 \\
    Co-CoOp  & 64.07 & 48.75 & 50.63 &76.18 & 59.91 \\
    Ours  & \textbf{64.34} & \textbf{48.83} & \textbf{50.80} & \textbf{76.68} & \textbf{60.16} \\ \bottomrule
    \end{tabular}
    \label{domain_generalization}
\end{table*}
\subsection{Experimental Settings}
\textbf{Datasets}
For base-to-new class generalization and cross-dataset evaluation, we follow \cite{zhouetal2022cocoop, zhouetal2022coop} and apply 11 datasets: ImageNet]\cite{Imagenetfeifei}, Caltech101\cite{caltech101}, OxfordPets\cite{oxfordpets}, StanfordCars\cite{stanfordcars}, Flowers102 \cite{flowers102}, Food101\cite{food}, FGVCAircraft\cite{aircraft}, SUN39\cite{sun397}, UCF101\cite{ucf101},DTD\cite{dtd}, and EuroSAT\cite{eurosat}. 
For domain generalization experiments, we use ImageNet\cite{Imagenetfeifei} as the source dataset and four ImageNet variants as target datasets, including ImageNetV2\cite{imagenetv2}, ImageNet-Sketch\cite{imagenetsketch}, ImageNet-A\cite{imageneta}, and ImageNet-R\cite{imagenetr}.
For text-image retrieval, we use MSCOCO\cite{mscoco} which contains 113K/5K/5K for train/validation/test, and Flickr30k\cite{Retrieval} which contains 29K/1K/1K for train/validation/test.
For Visual Question Answering, we use VQAv2\cite{VQA}, which contains images paired with questions and answers.

\textbf{Baselines}
The primary rival to our approach is CoCoOp\cite{zhouetal2022cocoop}, which learns a single, general prompt and a single instance specific bias in comparison with our chained prompts and step-specific biases. Another rival is  CoOp\cite{zhouetal2022coop}, which learns fixed, static prompts. We also compare our result with CLIP\cite{CLIP}, which uses carefully manually designed prompts that are tuned using all classes.

\subsection{Implementation Details}
Consistent with CoCoOp, we use the vision backbone ViT-B/16\cite{CLIP} in CLIP. In our experiments, we use a chain length of three and initialize each of them with the phrase "a photo of a" using the pre-trained word embeddings. In all three experiments of image classification, we use the class label as the class description. In image-text retrieval, we use the image captions as the class labels, and for visual question answering, we use the concatenation of each question and answer pair as class labels. We use a learning rate of 0.002 to train the model with a batch size 1 for 10 epochs. Our source code will be submitted in the supplemental materials.

\subsection{Results and Analysis}
\textbf{Base-to-New Generalization.}
The comparitive result for our experiment in base-to-new setting is demonstrated in Table \ref{table_1}. Our method outperforms CoCoOp, CoOp, and CLIP in Harmonic Mean score for all 11 datasets. Demonstrating an outstanding generalizating ability. On new class accuracy, our method also surpasses CoCoOp on all 11 datasets, demonstraing a significantly better generalization ability.

\textbf{Cross Dataset Evaluation.}
The comparative results of cross dataset transfer is shown in Table \ref{cross_dataset}. Our method attains the highest accuracy on eight out of 10 datasets, and outperforms CoOp on all 10 datasets, indicating a strong ability in cross dataset transfer.

\textbf{Domain Generalization.}
 The comparative results of domain generalization is demonstrated in Table \ref{domain_generalization}. We test the model trained on ImageNet on four other Imagenet datasets. On all four datasets, our method is the best-performing one, surpassing CoCoOp, CoCp, and ClIP. This greatly demonstrates our method's capacity in OOD settings.

\begin{table}
\centering
\footnotesize
\caption{The comparative results of using 0.5\%, 1.0\%, and 2.0\% of the training data on MSCOCO and Flickr30k datasets. Our method outperforms CoCoOp and CLIP in all cases.}
\label{image-text-retrieval}
\begin{tabular}{cccc} 
\toprule
Training Data&Method&Flickr30k&MSCOCO\\
\midrule
0\% & CLIP & 83.00 & 53.30  \\ \midrule
\multirow{2}{*}{0.5\%} & CoCoOp & 82.80 & 53.50  \\
&CoT & \textbf{83.50} & \textbf{55.80}  \\ \midrule
\multirow{2}{*}{1.0\%} & CoCoOp & 84.50 & 56.40  \\
&CoT & \textbf{85.10} & \textbf{56.70}  \\ \midrule
\multirow{2}{*}{2.0\%} & CoCoOp & 85.00 & 57.00  \\
& CoT & \textbf{86.00} & \textbf{57.90}  \\\bottomrule
\end{tabular}
\end{table}

\textbf{Image-Text Retrieval.}
In this experiment, we test on two image captioning datasets: Flickr30k and MSCOCO. We sample only $1\%, 1.5\%, 2\%$ data points from the training set to train our model and test it on the test set. 
The comparison results are shown in Table \ref{image-text-retrieval}. Our method surpasses the zero-shot CLIP and CoCoOp in all tests, indicating that our method is capable of handling tasks that require more reasoning.

\begin{table}
\centering
\caption{The comparative results of using 0.25\%, 0.5\%, and 0.75\% of the training data on VQAv2 dataset. Our method outperforms CoCoOp and CLIP in all cases.}
\footnotesize
\setlength\tabcolsep{16pt}
\begin{tabular}{ccc} 
\toprule
Training Data&Method&VQAv2\\
\midrule
0\% & CLIP &  11.83  \\ \midrule
\multirow{2}{*}{0.25\%} & CoCoOp & 27.23\\
&Ours & \textbf{29.13}  \\ \midrule
\multirow{2}{*}{0.5\%} & CoCoOp & 29.51\\
&Ours & \textbf{30.72}  \\ \midrule
\multirow{2}{*}{0.75\%} & CoCoOp & 30.76\\
& Ours & \textbf{30.86}   \\ \bottomrule
\end{tabular}
\label{VQA}
\end{table}

\textbf{Visual Question Answering.}
In this experiment, we sample $0.25\%, 0.5\%, 0.75\%$ from the training set and test on the test set. The results are shown in Table \ref{VQA}. Our method constantly outperforms CLIP and CoCoOp in all settings with large margins, demonstrating that our approach can be very beneficial in complicated visual tasks that require extensive reasoning.

 \textbf{Visualizing Chain of Thought Prompts}
 The chain of thought prompts in our architecture are latent vectors, which means they cannot be expressed explicitly in words. However, we could visualize the process in two ways: 1. Study the gradual properties of the prompts:
we take out the prompts from each step of the chain and compute its maximum similarity with the extracted image feature. We found that with the proceeding of the chain, the similarity score gets higher, indicating that the prompts are gradually getting more reasonable 2.Chain length: Comparing it with simply averaging prompts, we showed that the prompts are not simply a modular way to attain more information, but rather benefits from the step-to-step process.

\subsection{Ablation Study}

\begin{figure}
\centering
\includegraphics[width=8cm]{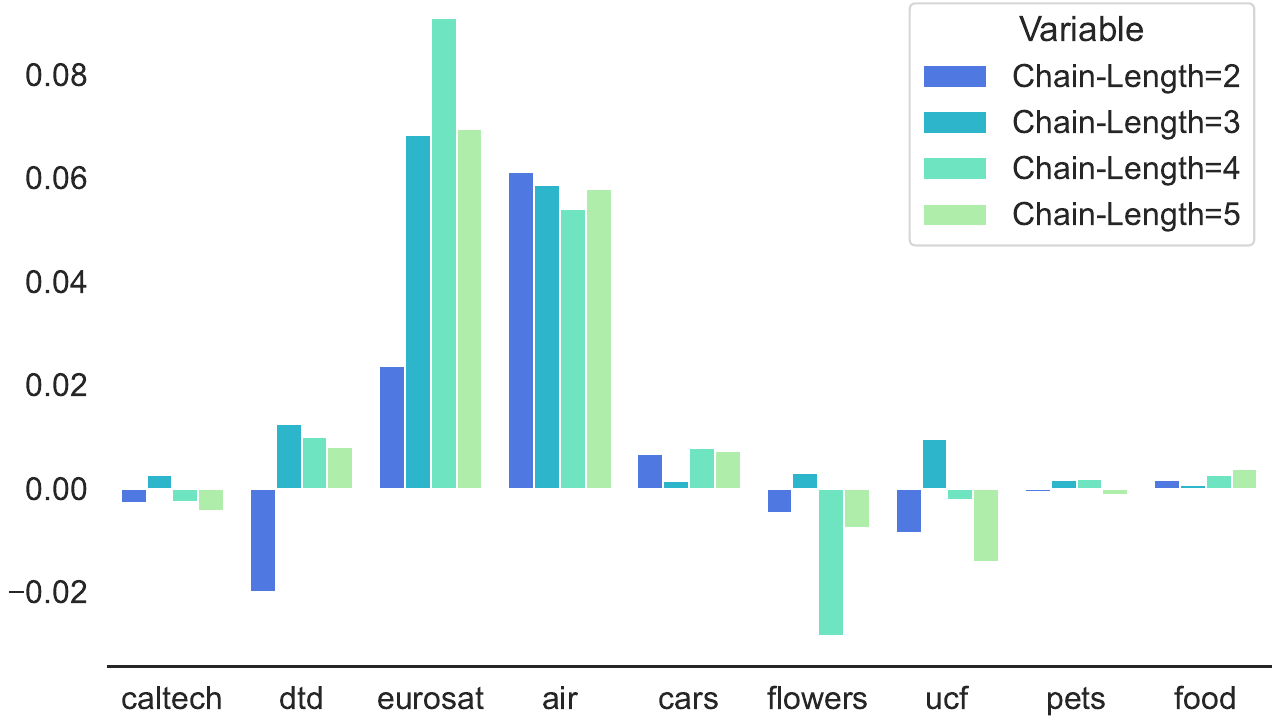}
\caption{The comparative results of using different chain lengths on 9 datasets. Short chain lengths melt down to the case of CoCoOp, while long chain lengths bring instability on Flowers and UCF tasks. The optimal choice of chain length is 3.}
\label{lengthfig}
\end{figure}

\textbf{Effect of chain length.}
In this study, we test the effect of different chain lengths. We conduct image classification base-to-new experiment with a chain length ranging from 2 to 5 (chain length = 1 is identical to CoCoOp). We compute the Harmonic Mean score and subtract it by the score of CoCoOp. The results are demonstrated in Figure \ref{lengthfig} and Table \ref{chainlen}. In general, the chain length of 3 attains the best average performance. Longer chains are less stable and more sensitive to the specific datasets, attaining significantly better results on some of the datasets but even falling behind CoCoOp on others. 
\begin{table*}
\centering
\setlength\tabcolsep{8pt}
\caption{The comparative results between using unchained Meta-Nets and using chained Meta-Nets. The chain length is set to 3 in both settings. We test the H score on 9 datasets. On 8 out of 9 datasets, the chained Meta-Nets outperforms the unchained Meta-nets.}
\footnotesize
    \begin{tabular}{ccccccccccc}
    \toprule
     Method&
    \rotatebox{90}{Caltech101}&
    \rotatebox{90}{DTD}&
    \rotatebox{90}{EuroSAT}&
    \rotatebox{90}{FGVCAircraft}&
    \rotatebox{90}{StanfordCars}&
    \rotatebox{90}{Flowers102}&
    \rotatebox{90}{UCF101}&
    \rotatebox{90}{OxfordPets}&
    \rotatebox{90}{Food101}&
    \rotatebox{90}{Average}\\ \midrule
    UnChained & 95.12 & 62.57& \textbf{78.37}& 33.13& 70.96& 81.54& 72.83& 96.27&
        90.80&75.73\\
    Chained & \textbf{96.11} & \textbf{66.11} & 78.61 & \textbf{33.61} & \textbf{72.17} & \textbf{82.01} & \textbf{78.06} & \textbf{96.59} & \textbf{91.07} & \textbf{77.15}\\ \bottomrule
    \end{tabular}
    \label{metanetchain}
\end{table*}

\begin{figure}
\centering
\includegraphics[width=8cm]{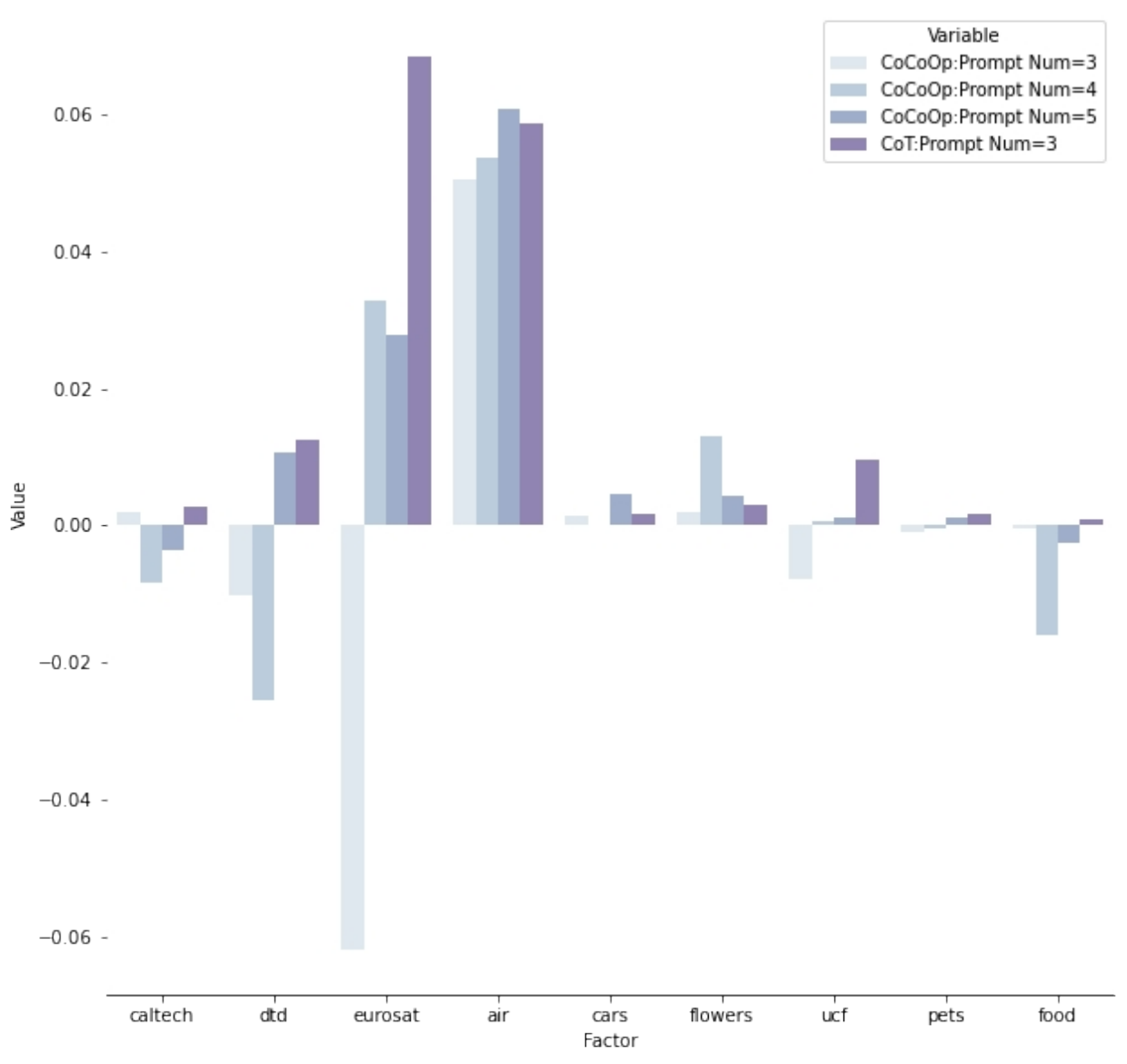}
\caption{The results of a heavier CoCoOp that takes average of more prompts and Meta-Nets compared with chained prompts and Meta-Nets. We compute the H score on 9 datasets. Our method even outperforms the heavier CoCoOp that contains 66.67\% more parameters than ours.}
\label{comparison}
\end{figure}

\textbf{Effect of the Chaining Architecture of Prompts.}
To validate that our performance gain is not attained by adding more parameters, we conduct experiments by adding more prompts and more Meta-Nets to the original CoCoOp architecture. Specifically, we train multiple prompts and multiple Meta-Nets with each Meta-Net generating a bias for one prompt; the only difference being that these prompts are not linked by a chain. Instead, they are parallel, and we simply took their average during training and prediction. The results are demonstrated in Figure \ref{comparison} and Table \ref{comp}. The chained architecture significantly outperforms the unchained architecture using the same number of parameters. It even outperforms the unchained architecture with 66.7\% more parameters. This result strongly demonstrates the superiority of our architectural design.

\textbf{Effect of the Chaining Architecture of Meta-Nets.}
In this study, we test the effect of linking the Meta-Nets into a chain. As a comparison, we merely cut off the chain between the Meta-Nets and let them learn their own biases. The results are shown in Table \ref{metanetchain}. We can see that adding a chain relationship could enhance the model's performance. Since the chain of thought requires step-by-step information, an information flow between Meta-Nets could lead to better performance.

\begin{figure}
\centering
\includegraphics[width=8cm]{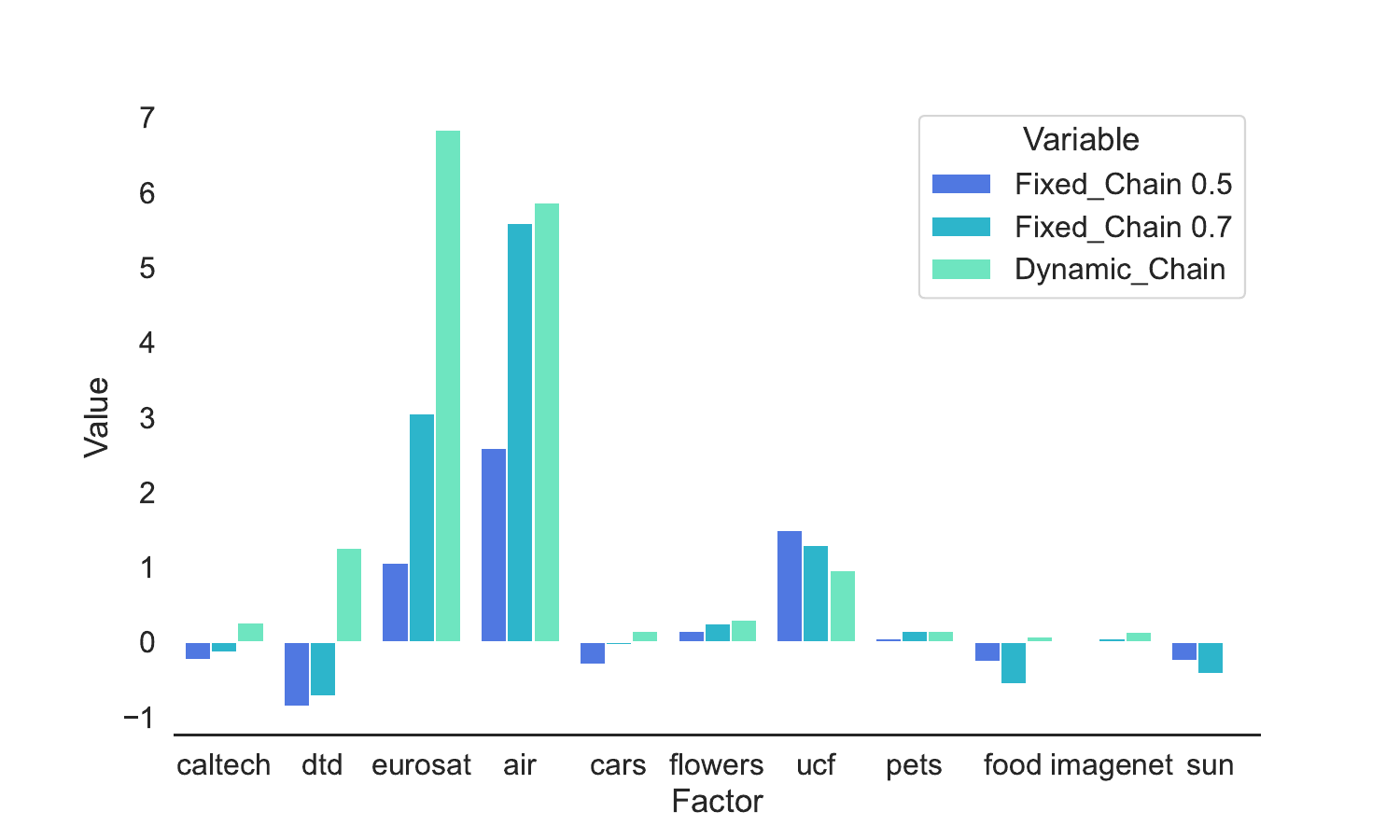}
\caption{A comparative result between using a fixed chain without a dynamic chain controller and using a dynamic chain controller. We conduct the base to new experiment on 9 datasets and assess the model performance with H scores. The dynamic chain controller attains
the best performance on 8 out 9 datasets.}
\label{weight}
\end{figure}

\textbf{Effect of the Dynamic Chain Controller.}
In our work, we apply a dynamic chain controller to control the chain based on the image input. To study the effect of this module, we conduct experiments where the parameters on the chain are fixed (we call this a fixed chain). Here, we demonstrate the results when using 0.5 and 0.7 as the parameters. More results are provided in the supplemental materials. The results are demonstrated in Figure \ref{weight} and Table \ref{dynamic_fig}. We can see that the dynamic chain controller performs much better. This is because different images prefer different reasoning processes, and dynamically control the chain based on the input could help the model adapt better to unseen classes.
\begin{table}
\setlength\tabcolsep{6pt}
\caption{The H score of taking the average of 3, 4, or 5 prompts and an according number of meta-nets in CoCoOp. Compared with using a chain structure with 3 prompts.}
\footnotesize
\centering
    \begin{tabular}{ccccc}
    \toprule
     &
    3-Average & 4-Average & 5-Average & 3-Chain \\ \midrule
    HM &  75.10 & 75.93 & 76.54 & \textbf{77.15}\\ \bottomrule
    \end{tabular}
    \label{comp}
\end{table}
\begin{table}
\caption{The average H score on 9 datasets with chain length of 2, 3, 4, and 5. Chain length of three is optimal in our experiment.}
\footnotesize
\label{ablation_dynamic_chain_controller}
\setlength\tabcolsep{8pt}
\centering
    \begin{tabular}{ccccc}
    \toprule
     &
    2-Chain & 3-Chain & 4-Chain & 5-Chain \\ \midrule
    HM &  76.01 & \textbf{77.15} & 76.87 & 76.70\\ \bottomrule
    \end{tabular}
    \label{chainlen}
\end{table}
\begin{table}
\setlength\tabcolsep{2pt}
\caption{The average H score of using a fixed chain with parameters 0.5 and 0.7 and using a dynamic chain controller.}
\footnotesize
\centering
    \begin{tabular}{cccc}
    \toprule
     &
    Fixed-Chain 0.5 & Fixed-Weight 0.7 & Dynamic Chain \\\midrule
    HM &  75.80 & 76.37 & \textbf{77.15}\\ \bottomrule
    \end{tabular}
    \label{dynamic_fig}
\end{table}


\section{Conclusion and Limitation}
In this work, we propose a Chain Of Thought prompt tuning framework that aims to model the step-to-step reasoning process that human conducts when solving a new task. We are the first to introduce this approach into vision domain and the first to design an effective framework to model this process, shedding new light in this field. We conduct extensive experiments including image-classification which consists of three sub-tasks, image-text retrieval using two datasets, and visual question answering using VQAv2 datasets. Experimental results demonstrate the effectiveness of our method. Besides, we conduct ablation experiments that demonstrate the superiority of our architectual design. In the future, we plan to design more sophisicated chain-of-thought prompt tuning methods and apply this intuition to other vision tasks. The primary limitation is the optimal number of prompts may vary over different downstream tasks and datasets. Yet we believe such limitation will not conceal the main contributions of our work and it could be released by adding a mechanism to choose the length of the chain based on the downstream tasks.


{\small
 \bibliographystyle{ieee_fullname}
 \bibliography{egbib}

\begin{thebibliography}{10}\itemsep=-1pt

\bibitem{inductivebias}
Y~Bengio A~Goyal.
\newblock Inductive biases for deep learning of higher-level cognition.
\newblock {\em Proceedings of the Royal Society A}, 2022.

\bibitem{gwt}
Alex Lamb Kartikeya Badola Nan Rosemary Ke Nasim Rahaman Jonathan Binas Charles
  Blundell Michael Mozer Yoshua~Bengio Anirudh~Goyal, Aniket~Didolkar.
\newblock Coordination among neural modules through a shared global workspace.
\newblock {\em ICLR}, 2022.

\bibitem{Zhupromptaligned}
Yucheng Han Yue~Wu Beier~Zhu, Yulei~Niu and Hanwang Zhang.
\newblock Prompt-aligned gradient for prompt tuning.
\newblock {\em arXiv preprint arXiv:2205.14865}, 2022.

\bibitem{imagenetv2}
Ludwig~Schmidt Benjamin~Recht, Rebecca~Roelofs and Vaishaal Shankar.
\newblock Do imagenet classifiers generalize to imagenet?
\newblock {\em ICML}, 2019.

\bibitem{Lesterprompt}
Rami Al-Rfou Brian~Lester and Noah Constant.
\newblock The power of scale for parameter-efficient prompt tuning.
\newblock {\em Conference on Empirical Methods in Natural Language Processing},
  2021.

\bibitem{Retrieval}
Chris M Cervantes Juan C Caicedo Julia~Hockenmaier Bryan A~Plummer, Liwei~Wang
  and Svetlana Lazebnik.
\newblock Flickr30k entities: Collecting region-to-phrase correspondences for
  richer imageto-sentence models.
\newblock {\em In Proceedings of the IEEE international conference on computer
  vision, pages 2641–2649.}, 2015.

\bibitem{ShoufaChen2022AdaptFormerAV}
Shoufa Chen, Chongjian Ge, Zhan Tong, Jiangliu Wang, Yibing Song, Jue Wang, and
  Ping Luo.
\newblock Adaptformer: Adapting vision transformers for scalable visual
  recognition.
\newblock 2022.

\bibitem{imagenetr}
Norman Mu Saurav Kadavath Frank Wang Evan Dorundo Rahul Desai Tyler Zhu Samyak
  Parajuli Mike Guo Dawn Song Jacob Steinhardt and Justin~Gilmer.
  Dan~Hendrycks, Steven~Basart.
\newblock The many faces of robustness: A critical analysis of
  out-of-distribution generalization.
\newblock {\em ICCV}, 2021.

\bibitem{imageneta}
Steven Basart Jacob~Steinhardt Dan~Hendrycks, Kevin~Zhao and Dawn Song.
\newblock Natural adversarial examples.
\newblock {\em CVPR}, 2021.

\bibitem{Flamingo}
Jean-Baptiste~Alayrac et al.
\newblock Flamingo: a visual language model for few-shot learning. arxiv
  preprint.
\newblock {\em arXiv:2204.14198}, 2022.

\bibitem{Gaoetal2021}
Tianyu Gao, Adam Fisch, and Danqi Chen.
\newblock Making pre-trained language models better few-shot learners.
\newblock {\em ArXiv}, abs/2012.15723, 2021.

\bibitem{imagenetsketch}
Zachary~Lipton Haohan~Wang, Songwei~Ge and Eric~P Xingr.
\newblock Learning robust global representations by penalizing local predictive
  power.
\newblock {\em NeurIPSL}, 2019.

\bibitem{humanreason}
Stephanie C. Y. Chan Antonia Creswell Dharshan Kumaran James L. McClelland
  Felix~Hill Ishita~Dasgupta, Andrew K.~Lampinen.
\newblock Language models show human-like content effects on reasoning.
\newblock {\em arXiv:2207.07051}, 2022.

\bibitem{Imagenetfeifei}
Richard Socher Li-Jia Li Kai~Li Jia~Deng, Wei~Dong and Li Fei-Fei.
\newblock Imagenet: A large-scale hierarchical image database.
\newblock {\em CVPR}, 2009.

\bibitem{sun397}
Krista A Ehinger-Aude~Oliva Jianxiong~Xiao, James~Hays and Antonio Torralba.
\newblock Sun database: Large-scale scene recognition from abbey to zoo.
\newblock {\em CVPR}, 2010.

\bibitem{stanfordcars}
Jia~Deng Jonathan~Krause, Michael~Stark and Li Fei-Fei.
\newblock 3d object representations for fine-grained categorization.
\newblock {\em ICCV-W}, 2013.

\bibitem{ucf101}
Amir Roshan~Zamir Khurram~Soomro and Mubarak Shah.
\newblock Ucf101: A dataset of 101 human actions classes from videos in the
  wild.
\newblock {\em arXiv preprint arXiv:1212.0402}, 2012.

\bibitem{Lietal2021}
Dongxu Li, Junnan Li, Hongdong Li, Juan~Carlos Niebles, and Steven C.~H. Hoi.
\newblock Align and prompt: Video-and-language pre-training with entity
  prompts.
\newblock {\em ArXiv}, abs/2112.09583, 2021.

\bibitem{caltech101}
Rob~Fergus Li~Fei-Fei and Pietro Perona.
\newblock Learning generative visual models from few training examples: An
  incremental bayesian approach tested on 101 object categories.
\newblock {\em CVPR-W}, 2004.

\bibitem{YuningLu2022PromptDL}
Yuning Lu, Jianzhuang Liu, Yonggang Zhang, Yajing Liu, and Xinmei Tian.
\newblock Prompt distribution learning.
\newblock 2022.

\bibitem{florence}
Yi-Ling Chen-Noel Codella Xiyang Dai Jianfeng Gao Houdong Hu Xuedong Huang
  Boxin Li Chunyuan Li Ce Liu Mengchen Liu Zicheng Liu Yumao Lu Yu Shi Lijuan
  Wang Jianfeng Wang Bin Xiao Zhen Xiao Jianwei Yang Michael Zeng Luowei Zhou
  Pengchuan~Zhang Lu~Yuan, Dongdong~Chen.
\newblock Florence: A new foundation model for computer vision.
\newblock {\em arXiv:2111.11432}, 2021.

\bibitem{food}
Matthieu~Guillaumin Lukas~Bossard and Luc~Van Gool.
\newblock Food-101–mining discriminative components with random forests.
\newblock {\em ECCV}, 2014.

\bibitem{dualreasoning}
Joshua B. Tenenbaum Brenden M.~Lake Maxwell~Nye, Michael Henry~Tessler.
\newblock Improving coherence and consistency in neural sequence models with
  dual-system, neuro-symbolic reasoning.
\newblock {\em NeurIPS}, 2021.

\bibitem{dtd}
Iasonas Kokkinos Sammy~Mohamed Mircea~Cimpoi, Subhransu~Maji and Andrea
  Vedaldi.
\newblock Describing textures in the wild.
\newblock {\em CVPR}, 2014.

\bibitem{flowers102}
Maria-Elena Nilsback and Andrew Zisserman.
\newblock Automated flower classification over a large number of classes.
\newblock {\em ICVGIP}, 2008.

\bibitem{oxfordpets}
Andrew~Zisserman Omkar M~Parkhi, Andrea~Vedaldi and CV Jawahar.
\newblock Cats and dogs.
\newblock {\em CVPR}, 2012.

\bibitem{learntoexplain}
Tony Xia Liang Qiu Kai-Wei Chang Song-Chun Zhu Oyvind Tafjord Peter Clark
  Ashwin~Kalyan Pan~Lu, Swaroop~Mishra.
\newblock Learn to explain: Multimodal reasoning via thought chains for science
  question answering.
\newblock {\em Neurips}, 2022.

\bibitem{eurosat}
Andreas~Dengel Patrick~Helber, Benjamin~Bischke and Damian Borth.
\newblock Eurosat: A novel dataset and deep learning benchmark for land use and
  land cover classification.
\newblock {\em IEEE Journal of Selected Topics in Applied Earth Observations
  and Remote Sensing}, 2019.

\bibitem{vqa-caption}
Chris Buehler Damien Teney Mark Johnson Stephen~Gould Peter~Anderson,
  Xiaodong~He and Lei Zhang.
\newblock Bottom-up and top-down attention for image captioning and visual
  question answering.
\newblock {\em CVPR}, 2018.

\bibitem{CLIP}
Alec Radford, Jong~Wook Kim, Chris Hallacy, Aditya Ramesh, Gabriel Goh,
  Sandhini Agarwal, Girish Sastry, Amanda Askell, Pamela Mishkin, Jack Clark,
  Gretchen Krueger, and Ilya Sutskever.
\newblock Learning transferable visual models from natural language
  supervision.
\newblock {\em international conference on machine learning}, 2021.

\bibitem{VQA}
Hao Tan Mohit Bansal Anna Rohrbach Kai-Wei Chang Zhewei~Yao Sheng~Shen, Liunian
  Harold~Li and Kurt Keutzer.
\newblock How much can clip benefit vision-and-language tasks?
\newblock {\em arXiv:2107.06383}, 2021.

\bibitem{TaylorShin2020AutoPromptEK}
Taylor Shin, Yasaman Razeghi, Robert~L. Logan, Eric Wallace, and Sameer Singh.
\newblock Autoprompt: Eliciting knowledge from language models with
  automatically generated prompts.
\newblock {\em empirical methods in natural language processing}, 2020.

\bibitem{aircraft}
Juho Kannala Matthew~Blaschko Subhransu~Maji, Esa~Rahtu and Andrea Vedaldi.
\newblock Fine-grained visual classification of aircraft.
\newblock {\em arXiv preprint arXiv:1306.5151}, 2013.

\bibitem{GPT3}
Nick Ryder Melanie Subbiah Jared Kaplan Prafulla Dhariwal Arvind Neelakantan
  Pranav Shyam Girish Sastry Amanda Askell Sandhini Agarwal Ariel Herbert-Voss
  Gretchen Krueger Tom Henighan Rewon Child Aditya Ramesh Daniel M. Ziegler
  Jeffrey Wu Clemens Winter Christopher Hesse Mark Chen Eric Sigler Mateusz
  Litwin Scott Gray Benjamin Chess Jack Clark Christopher Berner Sam McCandlish
  Alec Radford Ilya Sutskever Dario~Amodei Tom B.~Brown, Benjamin~Mann.
\newblock Language models are few-shot learners.
\newblock {\em arXiv:2005.14165}, 2020.

\bibitem{mscoco}
Serge Belongie James Hays Pietro Perona Deva Ramanan Piotr~Dollár
  Tsung-Yi~Lin, Michael~Maire and C~Lawrence Zitnick.
\newblock Microsoft coco: Common objects in context.
\newblock {\em ECCV}, 2014.

\bibitem{ZifengWang2022LearningTP}
Zifeng Wang, Zizhao Zhang, Chen-Yu Lee, Han Zhang, Ruoxi Sun, Xiaoqi Ren,
  Guolong Su, Vincent Perot, Jennifer Dy, and Tomas Pfister.
\newblock Learning to prompt for continual learning.
\newblock 2022.

\bibitem{JasonWei2022ChainOT}
Jason Wei, Xuezhi Wang, Dale Schuurmans, Maarten Bosma, Ed Chi, Quoc Le, and
  Denny Zhou.
\newblock Chain of thought prompting elicits reasoning in large language
  models.
\newblock 2022.

\bibitem{innermonologue}
Ted Xiao Harris Chan Jacky Liang Pete Florence Andy Zeng Jonathan Tompson Igor
  Mordatch Yevgen Chebotar Pierre Sermanet Noah Brown Tomas Jackson Linda Luu
  Sergey Levine Karol Hausman Brian~Ichter Wenlong~Huang, Fei~Xia.
\newblock Inner monologue: Embodied reasoning through planning with language
  models.
\newblock {\em arXiv:2207.05608}, 2022.

\bibitem{LeweiYao2022FILIPFI}
Lewei Yao, Runhui Huang, Lu Hou, Guansong Lu, Minzhe Niu, Hang Xu, Xiaodan
  Liang, Zhenguo Li, Xin Jiang, and Chunjing Xu.
\newblock Filip: Fine-grained interactive language-image pre-training.
\newblock 2022.

\bibitem{XiaohuaZhai2021LiTZT}
Xiaohua Zhai, Xiao Wang, Basil Mustafa, Andreas Steiner, Daniel Keysers,
  Alexander Kolesnikov, and Lucas Beyer.
\newblock Lit: Zero-shot transfer with locked-image text tuning.
\newblock {\em arXiv: Computer Vision and Pattern Recognition}, 2021.

\bibitem{XinZhang2021DomainPL}
Xin Zhang, Yusuke Iwasawa, Yutaka Matsuo, and Shixiang~Shane Gu.
\newblock Domain prompt learning for efficiently adapting clip to unseen
  domains.
\newblock 2021.

\bibitem{zhouetal2022cocoop}
Kaiyang Zhou, Jingkang Yang, Chen~Change Loy, and Ziwei Liu.
\newblock Conditional prompt learning for vision-language models.
\newblock In {\em IEEE/CVF Conference on Computer Vision and Pattern
  Recognition (CVPR)}, 2022.

\bibitem{zhouetal2022coop}
Kaiyang Zhou, Jingkang Yang, Chen~Change Loy, and Ziwei Liu.
\newblock Learning to prompt for vision-language models.
\newblock {\em International Journal of Computer Vision (IJCV)}, 2022.

\end{thebibliography}
 }
\newpage
\appendix
\begin{table*}
\centering
\setlength\tabcolsep{8pt}
\caption{The comparative result between using the final prompt, concatenating the last 2 prompts, and concatenating the last three prompts. We conduct the base to new experiment and use H score as the metric. Using the final prompt for prediction attains the highest average H score and attains the best performance in 6 out of 9 datasets. While using a concatenation of more prompts leads to a less stable performance. We assume that unlike NLP, feeding the entire chain of thought might not be as beneficial in vision domain.}
\footnotesize
    \begin{tabular}{ccccccccccc}
    \toprule
     Method&
    \rotatebox{90}{Caltech101}&
    \rotatebox{90}{DTD}&
    \rotatebox{90}{EuroSAT}&
    \rotatebox{90}{FGVCAircraft}&
    \rotatebox{90}{StanfordCars}&
    \rotatebox{90}{Flowers102}&
    \rotatebox{90}{UCF101}&
    \rotatebox{90}{OxfordPets}&
    \rotatebox{90}{Food101}&
    \rotatebox{90}{Average}\\ \midrule
    Single& 96.11 & \textbf{66.11} & \textbf{78.61} & 33.61 & 72.17 & \textbf{82.01} & \textbf{78.06} & \textbf{96.59} & \textbf{91.07} & \textbf{77.15}\\
    Concat 2 & 95.29&61.40&63.14&\textbf{35.46}&\textbf{72.75}&80.43&77.48&96.55&91.04&74.84\\
    Concat 3 & \textbf{96.29}&59.64&70.10&34.30&70.83&81.69&76.01&96.25&90.46&75.06\\
    \bottomrule
    \end{tabular}
    \label{concatenation}
\end{table*}
\begin{table*}
\centering
\setlength\tabcolsep{8pt}
\caption{The comparative results between using different backbones. We conduct the base to new experiment on 9 datasets and use the H score as the metric. Our results show that using different backbones could have a great impact on the model's performance, suggesting that finding a proper model backbone remains important.}
\footnotesize
    \begin{tabular}{ccccccccccc}
    \toprule
     Method&
    \rotatebox{90}{Caltech101}&
    \rotatebox{90}{DTD}&
    \rotatebox{90}{EuroSAT}&
    \rotatebox{90}{FGVCAircraft}&
    \rotatebox{90}{StanfordCars}&
    \rotatebox{90}{Flowers102}&
    \rotatebox{90}{UCF101}&
    \rotatebox{90}{OxfordPets}&
    \rotatebox{90}{Food101}&
    \rotatebox{90}{Average}\\ \midrule
    ViT-B/16& \textbf{96.11} & \textbf{66.11} & \textbf{78.61} & \textbf{33.61} & \textbf{72.17} & \textbf{82.01} & \textbf{78.06} & \textbf{96.59} & \textbf{91.07} & \textbf{77.15}\\
    ViT-B/32 & 95.10 & 61.71 &71.84 & 26.60 & 66.83 & 73.66 &
73.39 & 95.08 & 87.07 & 72.36\\
    RN50 & 93.45 & 58.88 & 49.44 & 24.78 & 63.95 & 75.51& 69.06&94.43&84.75&68.25\\
    RN101&94.42&62.14&54.51&7.35&69.28&75.41&72.45&94.29
&86.99&68.54\\
    \bottomrule
    \end{tabular}
    \label{backbones}
\end{table*}
\begin{table*}
\centering
\setlength\tabcolsep{8pt}
\caption{The comparative confidence between CoCoOp and our method. We conduct the base to new experiment on 9 datasets. We calculate the average confidence score (which is the max probability among all classes) over the instances that the model predicts correctly. For this metric, we want the value to be as high as possible so that the model is confident when it predicts correctly. Our method outperforms CoCoOp on 7 out of 9 datasets, demonstrating that our method is overall more confident in its correct choices.}
\footnotesize
    \begin{tabular}{ccccccccccc}
    \toprule
     Method&
    \rotatebox{90}{Caltech101}&
    \rotatebox{90}{DTD}&
    \rotatebox{90}{EuroSAT}&
    \rotatebox{90}{FGVCAircraft}&
    \rotatebox{90}{StanfordCars}&
    \rotatebox{90}{Flowers102}&
    \rotatebox{90}{UCF101}&
    \rotatebox{90}{OxfordPets}&
    \rotatebox{90}{Food101}&
    \rotatebox{90}{Average}\\ \midrule
    CoCoOp & 0.97&0.75&0.68&\textbf{0.61}&0.82&\textbf{0.91}&0.86&0.96&0.92&0.83\\
    Ours&\textbf{0.97}&\textbf{0.78}&\textbf{0.74}&0.59&\textbf{0.83}&0.90&\textbf{0.87}&\textbf{0.97}&\textbf{0.92}&\textbf{0.84}\\
    \bottomrule
    \end{tabular}
    \label{correct-confidence}
\end{table*}
\begin{table*}
\centering
\setlength\tabcolsep{8pt}
\caption{The comparative confidence between CoCoOp and our method. We conduct the base to new experiment on 9 datasets. We calculate the average confidence score (which is the max probability among all classes) over the instances that the model predicts wrong. For this metric, we want the value to be as low as possible so that the model is not so confident when it predicts wrong. Our method outperforms CoCoOp on 7 out of 9 datasets, demonstrating that our method is overall more 
hesitivein its wrong choices.}
\footnotesize
    \begin{tabular}{ccccccccccc}
    \toprule
     Method&
    \rotatebox{90}{Caltech101}&
    \rotatebox{90}{DTD}&
    \rotatebox{90}{EuroSAT}&
    \rotatebox{90}{FGVCAircraft}&
    \rotatebox{90}{StanfordCars}&
    \rotatebox{90}{Flowers102}&
    \rotatebox{90}{UCF101}&
    \rotatebox{90}{OxfordPets}&
    \rotatebox{90}{Food101}&
    \rotatebox{90}{Average}\\ \midrule
    CoCoOp & 0.69&\textbf{0.52}&0.57&0.35&0.55&0.63&0.60&\textbf{0.63}&0.59&0.57\\
    Ours&\textbf{0.68}&0.54&\textbf{0.53}&\textbf{0.35}&\textbf{0.54}&\textbf{0.60}&\textbf{0.57}&0.67&\textbf{0.59}&\textbf{0.56}\\
    \bottomrule
    \end{tabular}
    \label{wrong-confidence}
\end{table*}
\begin{table*}
\setlength\tabcolsep{6pt}
\caption{The implementation details with more hyperparameters.}
\footnotesize
\centering
    \begin{tabular}{ccccc}
    \toprule
     &
    Learning Rate & Prompt Length & Meta-Net Linear dim & Chain Controller dim\\ \midrule
    CoCoOp& 0.002 & 4 & dim // 16 & None\\
    Ours&  0.002 & 4 & dim // 16 & dim // 16\\ \bottomrule
    
    \end{tabular}
    \label{comp}
\end{table*}



\section{Supplementary Experimental Analysis}

\begin{table*}[!h]
\centering
\setlength\tabcolsep{8pt}
\caption{The comparative results between using different backbones. We conduct the base to new experiment on 9 datasets and use the H score as the metric. Our results show that using different backbones could have a great impact on the model's performance, suggesting that finding a proper model backbone remains important.}
\footnotesize
    \begin{tabular}{ccccccccccc}
    \toprule
     Method&
    \rotatebox{90}{Caltech101}&
    \rotatebox{90}{DTD}&
    \rotatebox{90}{EuroSAT}&
    \rotatebox{90}{FGVCAircraft}&
    \rotatebox{90}{StanfordCars}&
    \rotatebox{90}{Flowers102}&
    \rotatebox{90}{UCF101}&
    \rotatebox{90}{OxfordPets}&
    \rotatebox{90}{Food101}&
    \rotatebox{90}{Average}\\ \midrule
    ViT-B/16& \textbf{96.11} & \textbf{66.11} & \textbf{78.61} & \textbf{33.61} & \textbf{72.17} & \textbf{82.01} & \textbf{78.06} & \textbf{96.59} & \textbf{91.07} & \textbf{77.15}\\
    ViT-B/32 & 95.10 & 61.71 &71.84 & 26.60 & 66.83 & 73.66 &
73.39 & 95.08 & 87.07 & 72.36\\
    ResNet-50 & 93.45 & 58.88 & 49.44 & 24.78 & 63.95 & 75.51& 69.06&94.43&84.75&68.25\\
    ResNet-101&94.42&62.14&54.51&7.35&69.28&75.41&72.45&94.29
&86.99&68.54\\
    \bottomrule
    \end{tabular}
    \label{backbones}
\end{table*}

\begin{table*}[!h]
\centering
\setlength\tabcolsep{8pt}
\caption{The comparative result between using the final prompt, concatenating the last 2 prompts, and concatenating the last three prompts. We conduct the base to new experiment and use H score as the metric. Using the final prompt for prediction attains the highest average H score and attains the best performance in 6 out of 9 datasets. While using a concatenation of more prompts leads to a less stable performance. We assume that unlike NLP, feeding the entire chain of thought might not be as beneficial in vision domain.}
\footnotesize
    \begin{tabular}{ccccccccccc}
    \toprule
     Method&
    \rotatebox{90}{Caltech101}&
    \rotatebox{90}{DTD}&
    \rotatebox{90}{EuroSAT}&
    \rotatebox{90}{FGVCAircraft}&
    \rotatebox{90}{StanfordCars}&
    \rotatebox{90}{Flowers102}&
    \rotatebox{90}{UCF101}&
    \rotatebox{90}{OxfordPets}&
    \rotatebox{90}{Food101}&
    \rotatebox{90}{Average}\\ \midrule
    Single& 96.11 & \textbf{66.11} & \textbf{78.61} & 33.61 & 72.17 & \textbf{82.01} & \textbf{78.06} & \textbf{96.59} & \textbf{91.07} & \textbf{77.15}\\
    Concat 2 & 95.29&61.40&63.14&\textbf{35.46}&\textbf{72.75}&80.43&77.48&96.55&91.04&74.84\\
    Concat 3 & \textbf{96.29}&59.64&70.10&34.30&70.83&81.69&76.01&96.25&90.46&75.06\\
    \bottomrule
    \end{tabular}
    \label{concatenation}
\end{table*}

\begin{table*}[!h]
\centering
\setlength\tabcolsep{8pt}
\caption{The ablation study of using different prompt lenths. We conduct the base to new experiment on 9 datasets and use the H score as the metric. We use a prompt length of 4, 8 and 16 with random initialization. Our results show that longer prompt length with merely random initialization could attain promising results. Suggesting that finding a general initialization pattern for longer prompts lengths could be beneficial.}
\footnotesize
    \begin{tabular}{ccccccccccc}
    \toprule
     Method&
    \rotatebox{90}{Caltech101}&
    \rotatebox{90}{DTD}&
    \rotatebox{90}{EuroSAT}&
    \rotatebox{90}{FGVCAircraft}&
    \rotatebox{90}{StanfordCars}&
    \rotatebox{90}{Flowers102}&
    \rotatebox{90}{UCF101}&
    \rotatebox{90}{OxfordPets}&
    \rotatebox{90}{Food101}&
    \rotatebox{90}{Average}\\ \midrule
    Prompt len=4& 96.50&\textbf{66.12}&
68.54&
33.52&
71.74&
82.24&
76.73&
\textbf{96.69}&
\textbf{91.12}&\textbf{75.91}\\
    Prompt len=8 & \textbf{96.73}&64.81&69.52&33.15&71.76&82.46&\textbf{77.07}&
96.52&91.05&75.90\\
    Prompt len=16&95.84&58.75&\textbf{77.21}&\textbf{34.24}&\textbf{72.32}&\textbf{82.59}&74.74&96.47&90.97&75.90\\
    \bottomrule
    \end{tabular}
    \label{promptlen}
\end{table*}

\begin{table*}[!h]
\setlength\tabcolsep{6pt}
\caption{Some additional hyperparameters of our implementation. We use a learning rate of 0.002, a prompt length of 4 for each prompt. Denoting the output dimension from the image encoder as dim, we set the first linear layer's outut dim in Meta-Net as dim/16. We also set the first linear layer's output dim in the dynamic chain controller as dim/16.}
\footnotesize
\centering
\setlength\tabcolsep{14pt}
    \begin{tabular}{ccccc}
    \toprule
     &
    Learning Rate & Prompt Length & Meta-Net Linear dim & Chain Controller dim\\ \midrule
    CoCoOp& 0.002 & 4 & dim / 16 & None\\
    Ours&  0.002 & 4 & dim / 16 & dim / 16\\ \bottomrule
    \end{tabular}
    \label{implementation}
\end{table*}

\subsection{Concatenation of Prompts}

In NLP, the chain of thought inputs the full reasoning process into the language model, however in our case, we let the model learn the reasoning process and utilize the last decisive step to predict. In this experiment, we want to see if providing a concatenation of more prompts from previous phases may help with a vision task as well. In the base to new experiment, we test the concatenation of the last two prompts and the concatenation of all three prompts over 9 datasets.

The results are shown in Table \ref{concatenation}. A concatenation of additional prompts may improve performance on some datasets, like FGVCAircraft and StanfordCars while decreasing performance on others. As a result, feeding a very lengthy chain may be less advantageous for vision tasks than it is for NLP tasks.

\subsection{Effect of the Model BackBone}

In the previous experiments, to train and test our model, we employ the most generally used backbone, the ViT-B/16. We investigate the influence of three distinct pre-trained backbones in this experiment: ViT-B/32, RN50, and RN101. The results are presented in Table \ref{backbones}. The results show that the architecture chosen has a significant impact on the model's performance. Building a pre-trained vision-language model with improved adaptive capabilities is still worth paying attention to.

\subsection{Effect of the Prompt Length}
In this experiment, we conduct the ablation where different prompt lengths are used. Specifically, we use a prompt length of 8 with random initialization and prompt length of 16 with random initialization. The results are presented in Table \ref{promptlen}. Our results show thata longer prompt length with merely random initialization could attain promising results. Suggesting that finding a general initialization pattern for longer prompts lengths could be beneficial.
\section{Additional Visualization results}

\subsection{Model's Confidence Visualization}

\begin{figure}[htbp]
	\centering
	\begin{subfigure}{0.32\linewidth}
		\centering
		\includegraphics[width=0.9\linewidth,height=0.9\linewidth]{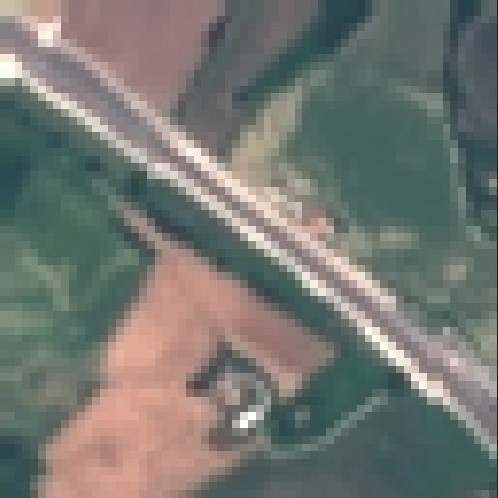}
		\caption{{Eurosat\\CoCoOp: 0.68\\\textbf{Ours: 0.74}}\centering}
		\label{Eurosat}
	\end{subfigure}
	\centering
	\begin{subfigure}{0.32\linewidth}
		\centering
		\includegraphics[width=0.9\linewidth,height=0.9\linewidth]{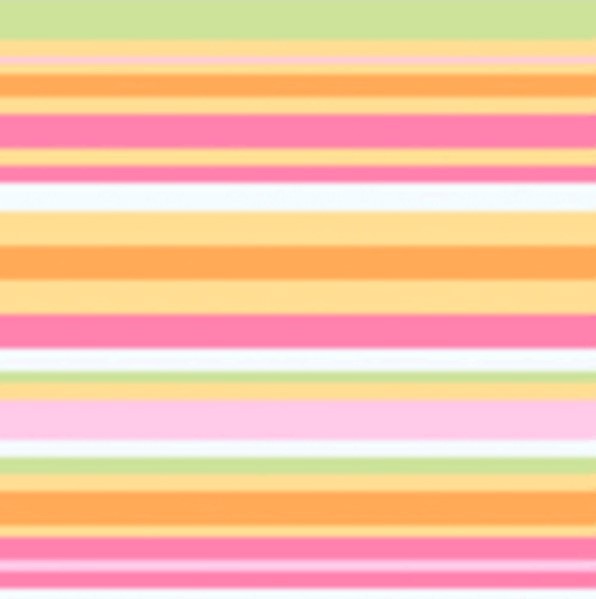}
		\caption{{DTD\\{CoCoOp: 0.75}\\\textbf{Ours: 0.80}}\centering}
		\label{DTD}
	\end{subfigure}
	\centering
	\begin{subfigure}{0.32\linewidth}
		\centering
		\includegraphics[width=0.9\linewidth,height=0.9\linewidth]{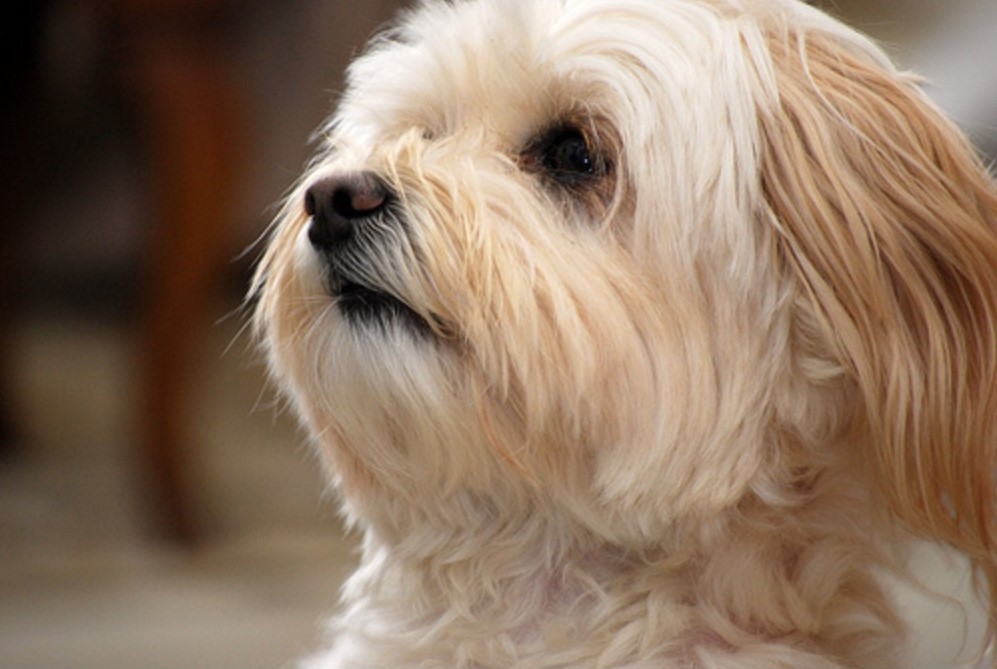}
		\caption{{Oxford Pets\\CoCoOp: 0.96\\\textbf{Ours: 0.97}}\centering}
		\label{Oxford Pets}
	\end{subfigure}
	
	\begin{subfigure}{0.32\linewidth}
		\centering
		\includegraphics[width=0.9\linewidth,height=0.9\linewidth]{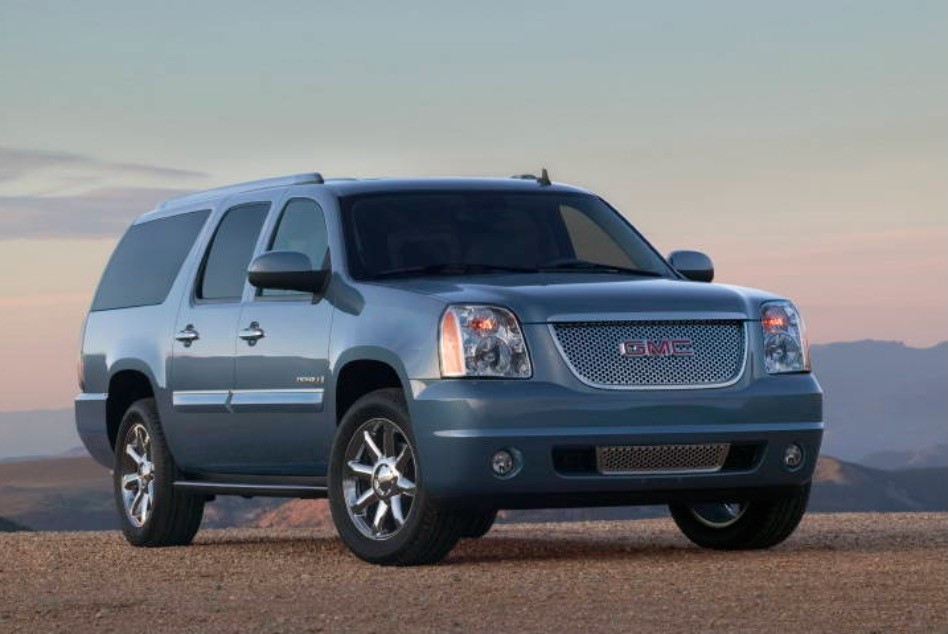}
		\caption{{Stanford Cars\\CoCoOp: 0.82\\\textbf{Ours: 0.83}}\centering}
		\label{Stanford_Cars}
	\end{subfigure}
	\centering
	\begin{subfigure}{0.32\linewidth}
		\centering
		\includegraphics[width=0.9\linewidth,height=0.9\linewidth]{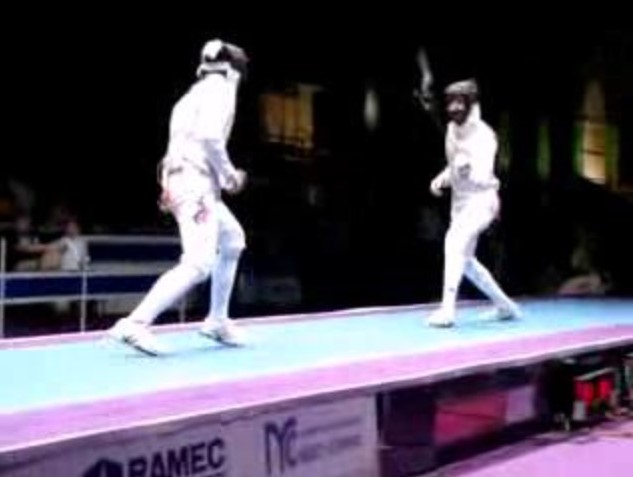}
		\caption{{UCF\\CoCoOp: 0.84\\\textbf{Ours: 0.87}}\centering}
		\label{UCF}
	\end{subfigure}
	\centering
	\begin{subfigure}{0.32\linewidth}
		\centering
		\includegraphics[width=0.9\linewidth,height=0.9\linewidth]{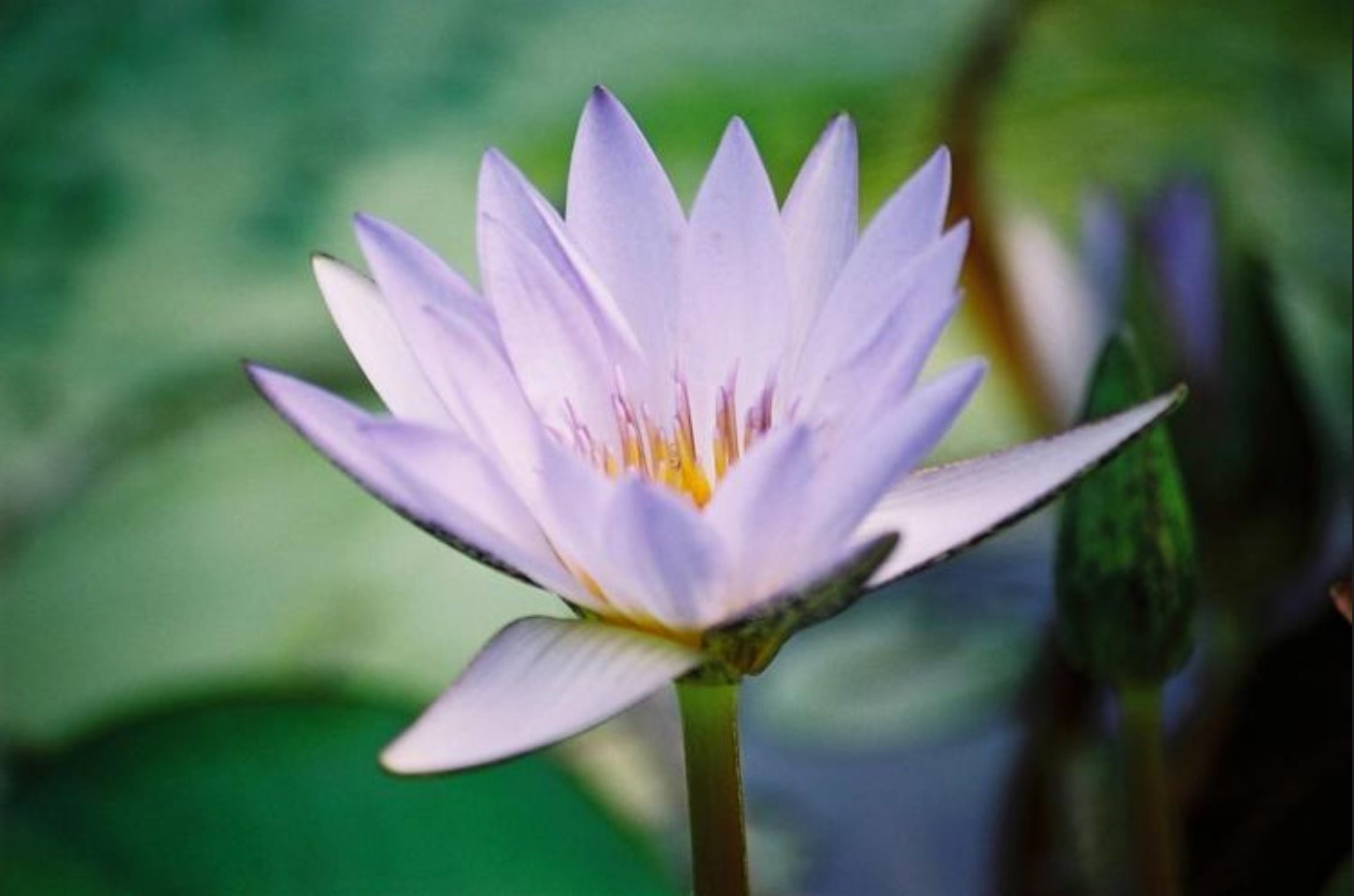}
		\caption{{Flowers\\CoCoOp: 0.86\\\textbf{Ours: 0.90}}\centering}
		\label{Flowers}
	\end{subfigure}
	\caption{Prediction confidence of our method compared with CoCoOp. We use the largest probability over all classes to indicate the model's confidence of that image. Our chain of thought prompting method is more confident than CoCoOp when giving predictions.}
	\label{figure_1}
\end{figure}

In this experiment, we want to test the confidence of the model. For each image, we use the largest probability value among all classes to demonstrate the model's confidence in its choice. We used the base to new experiment on 9 datasets. Figure \ref{figure_1} presents the models confidence. Compared with an unchained architecture, our model is more confident in its choices, indicating that our model might be able to better discriminate whether a choice is confidential or not.

\begin{figure*}[!h]
     \centering
     \begin{subfigure}[b]{0.3\textwidth}
         \centering
         \includegraphics[width=\textwidth]{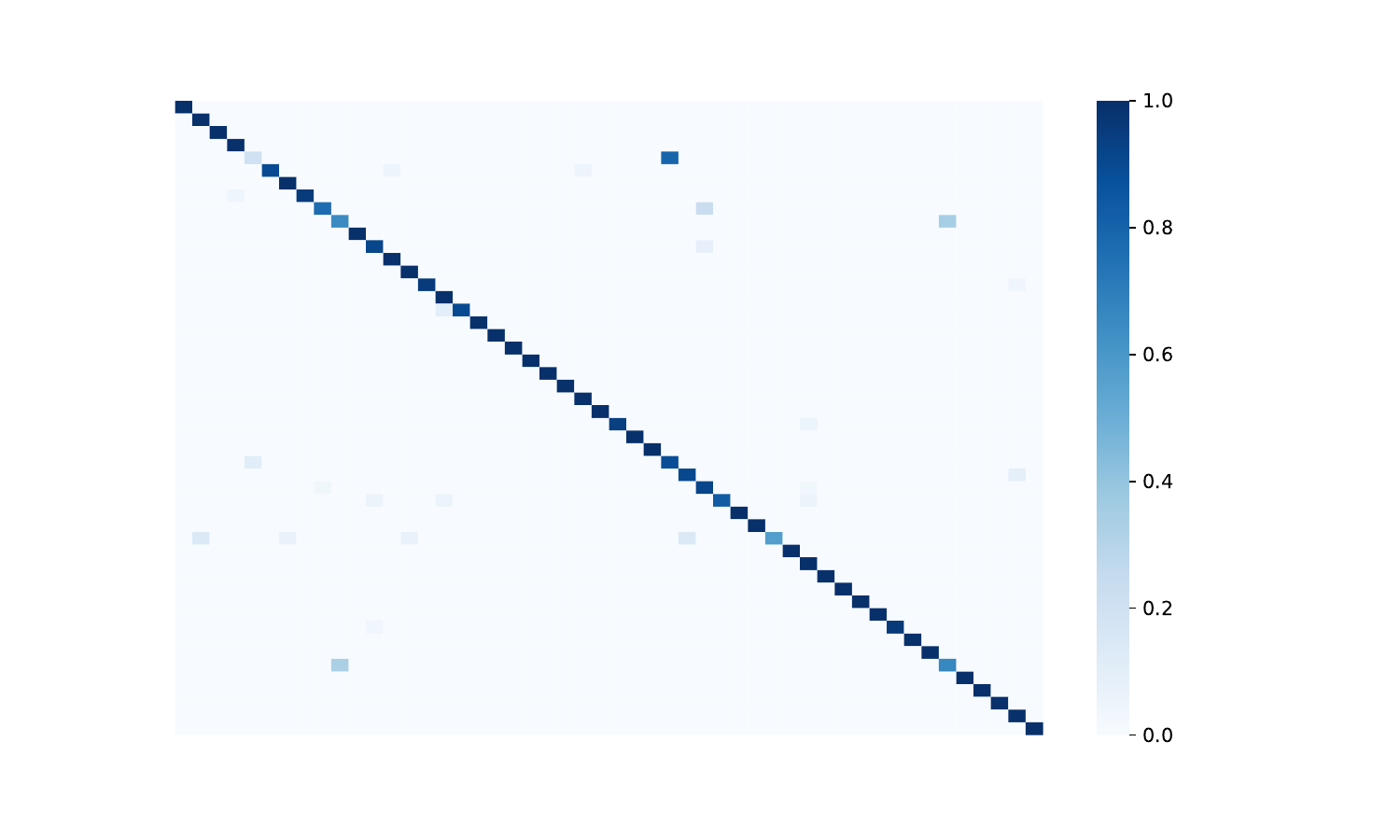}
         \caption{cocoop caltech101}
         \label{fig:y equals x}
     \end{subfigure}
     \begin{subfigure}[b]{0.3\textwidth}
         \centering
         \includegraphics[width=\textwidth]{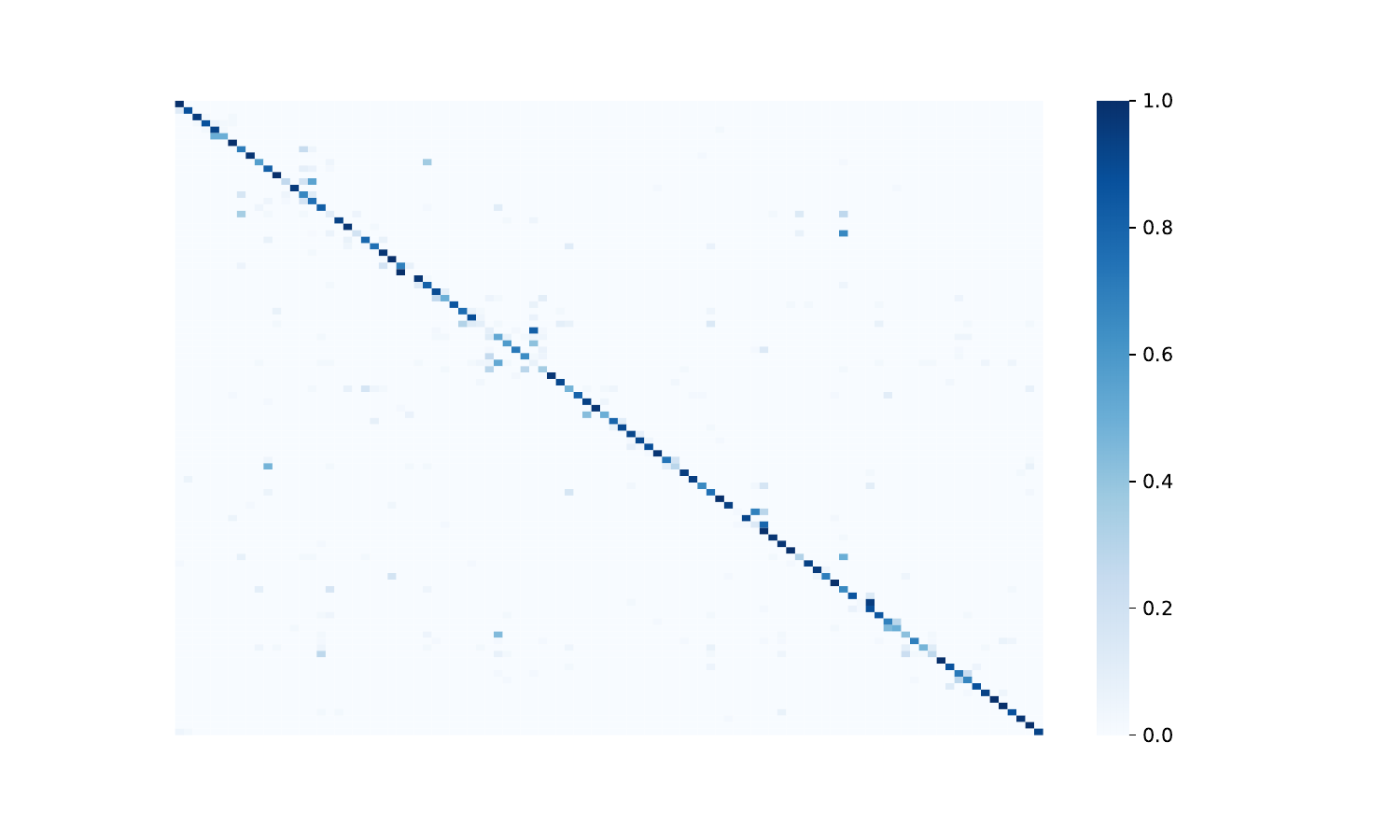}
         \caption{cocoop cars}
         \label{fig:three sin x}
     \end{subfigure}
     \begin{subfigure}[b]{0.3\textwidth}
         \centering
         \includegraphics[width=\textwidth]{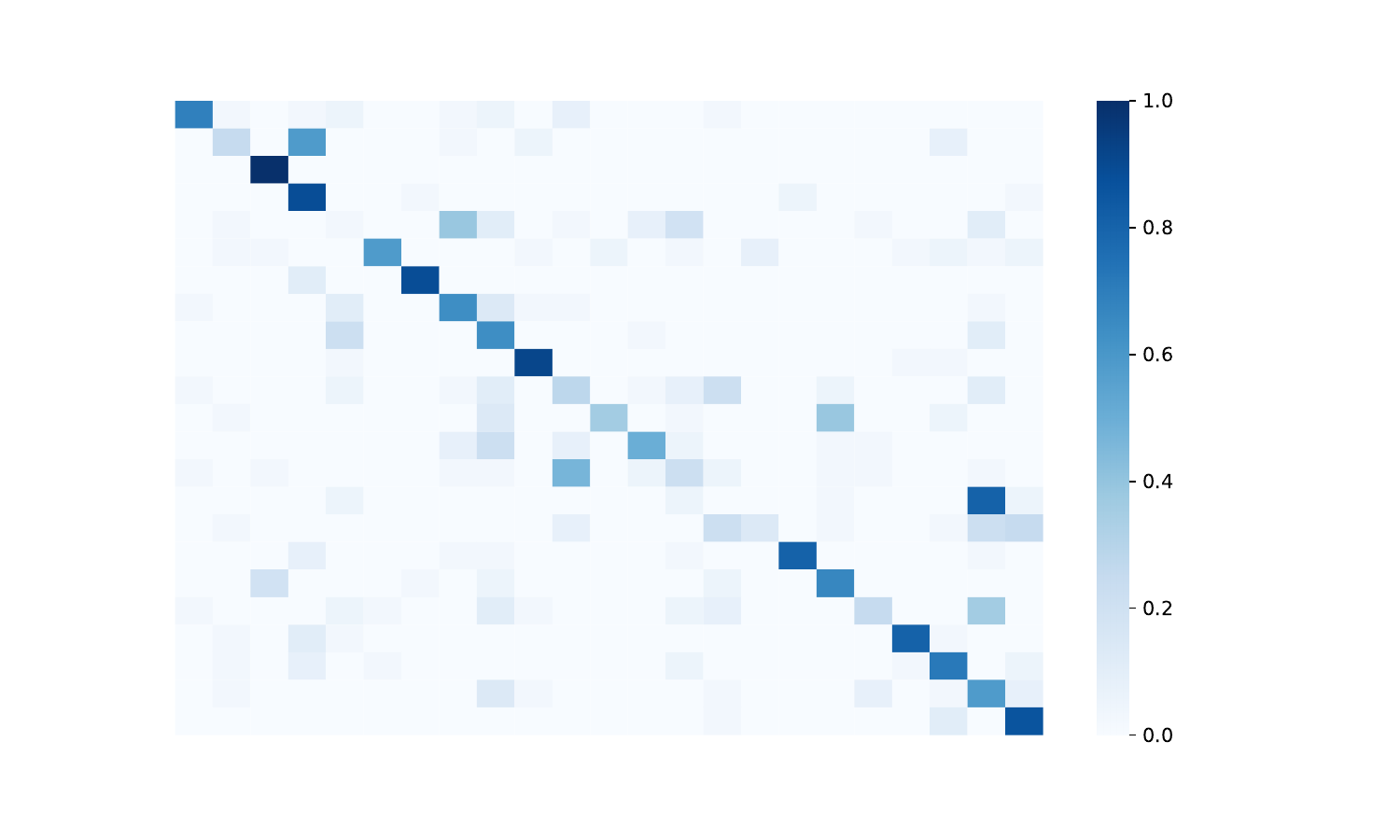}
         \caption{cocoop dtd}
         \label{fig:five over x}
     \end{subfigure}
     \begin{subfigure}[b]{0.3\textwidth}
         \centering
         \includegraphics[width=\textwidth]{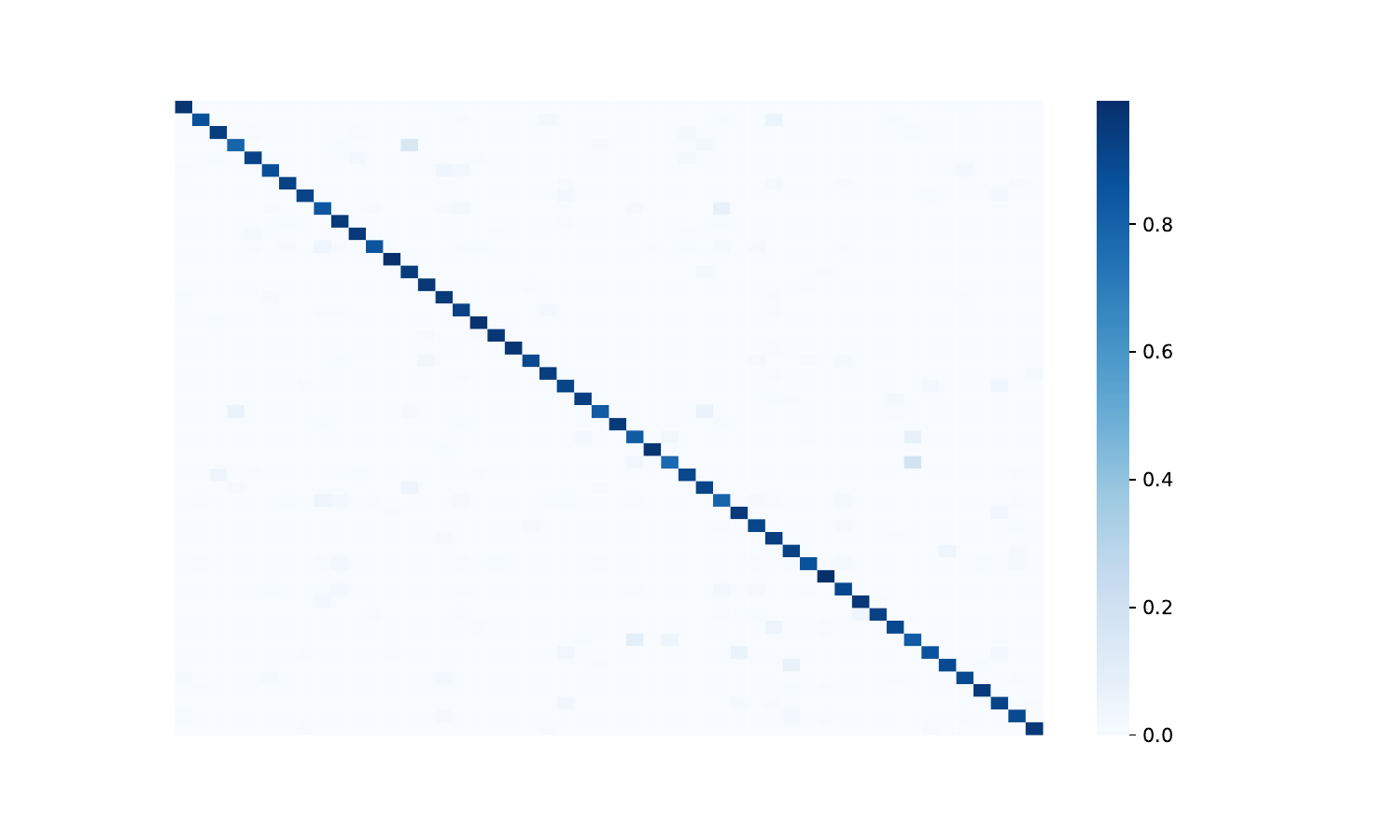}
         \caption{cocoop food}
         \label{fig:five over x}
     \end{subfigure}
     \begin{subfigure}[b]{0.3\textwidth}
         \centering
         \includegraphics[width=\textwidth]{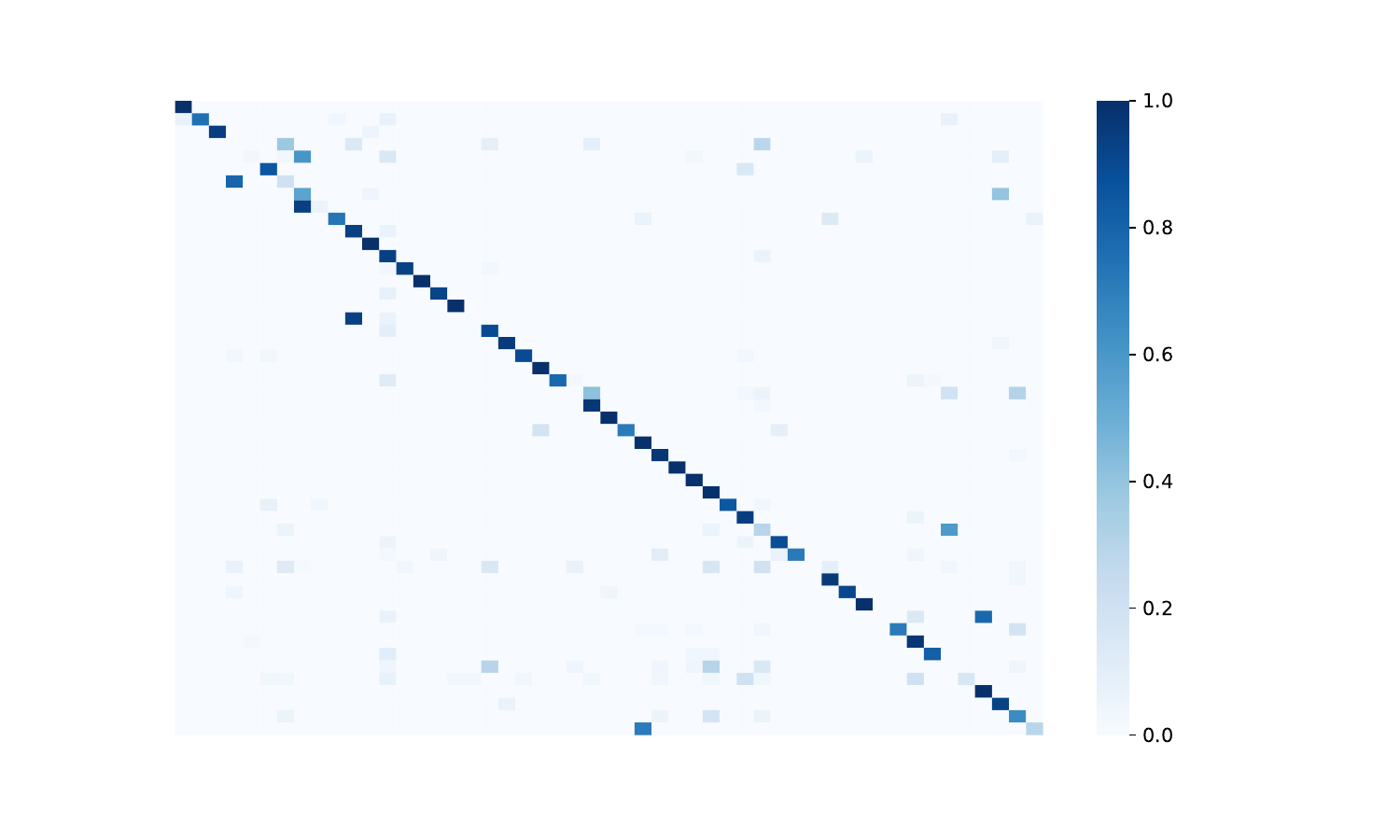}
         \caption{cocoop flowers}
         \label{fig:five over x}
     \end{subfigure}
     \begin{subfigure}[b]{0.3\textwidth}
         \centering
         \includegraphics[width=\textwidth]{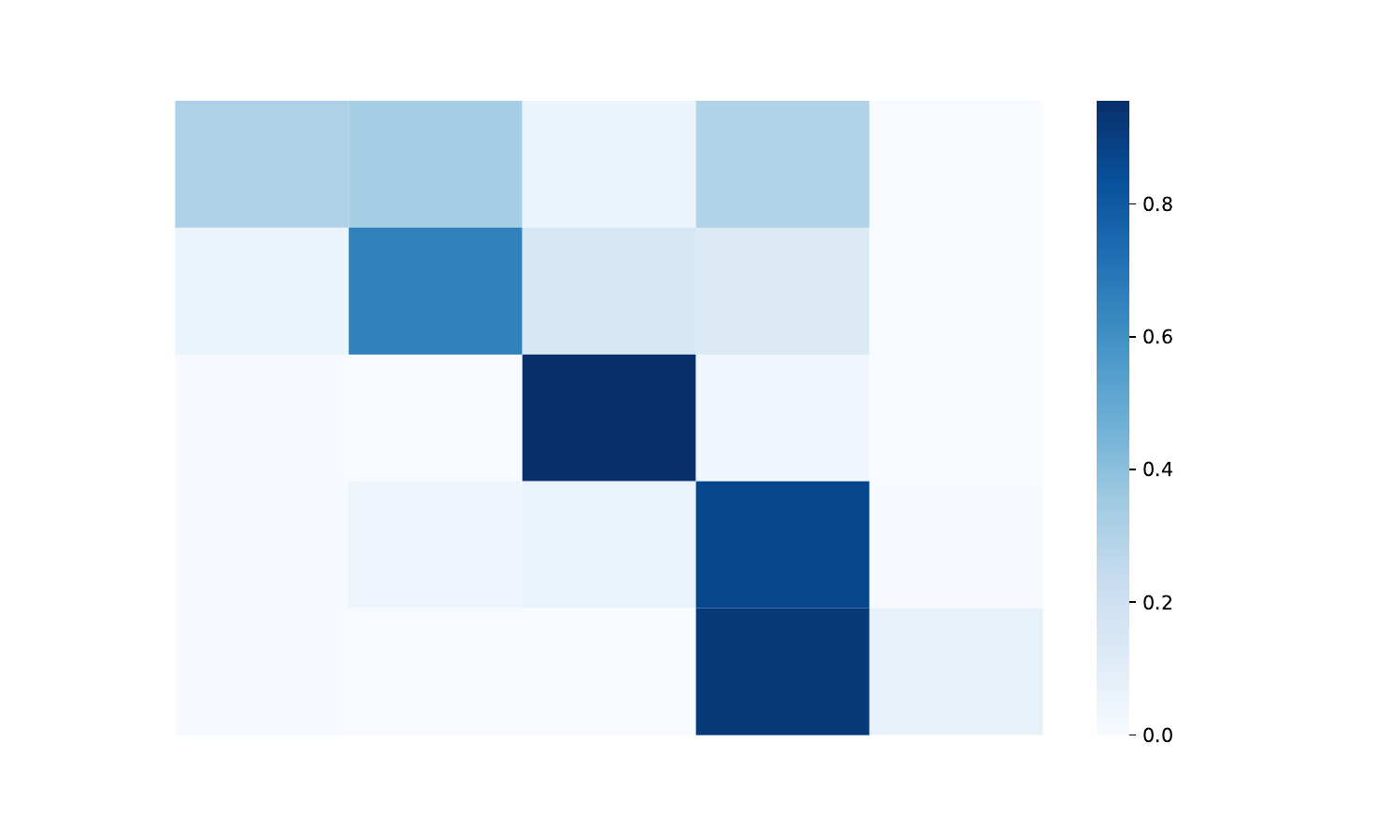}
         \caption{cocoop eurosat}
         \label{fig:five over x}
     \end{subfigure}
     \begin{subfigure}[b]{0.3\textwidth}
         \centering
         \includegraphics[width=\textwidth]{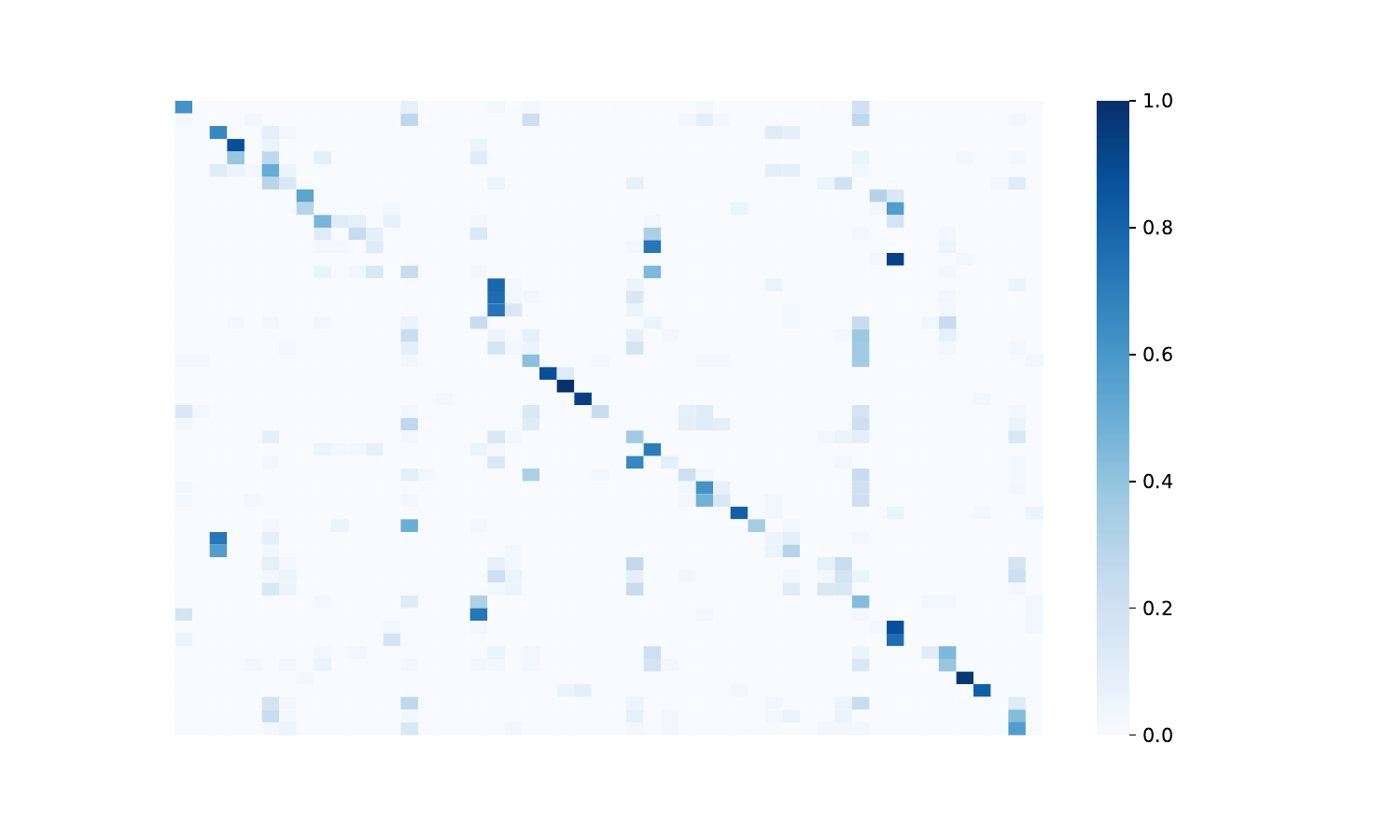}
         \caption{cocoop aircraft}
         \label{fig:five over x}
     \end{subfigure}
     \begin{subfigure}[b]{0.3\textwidth}
         \centering
         \includegraphics[width=\textwidth]{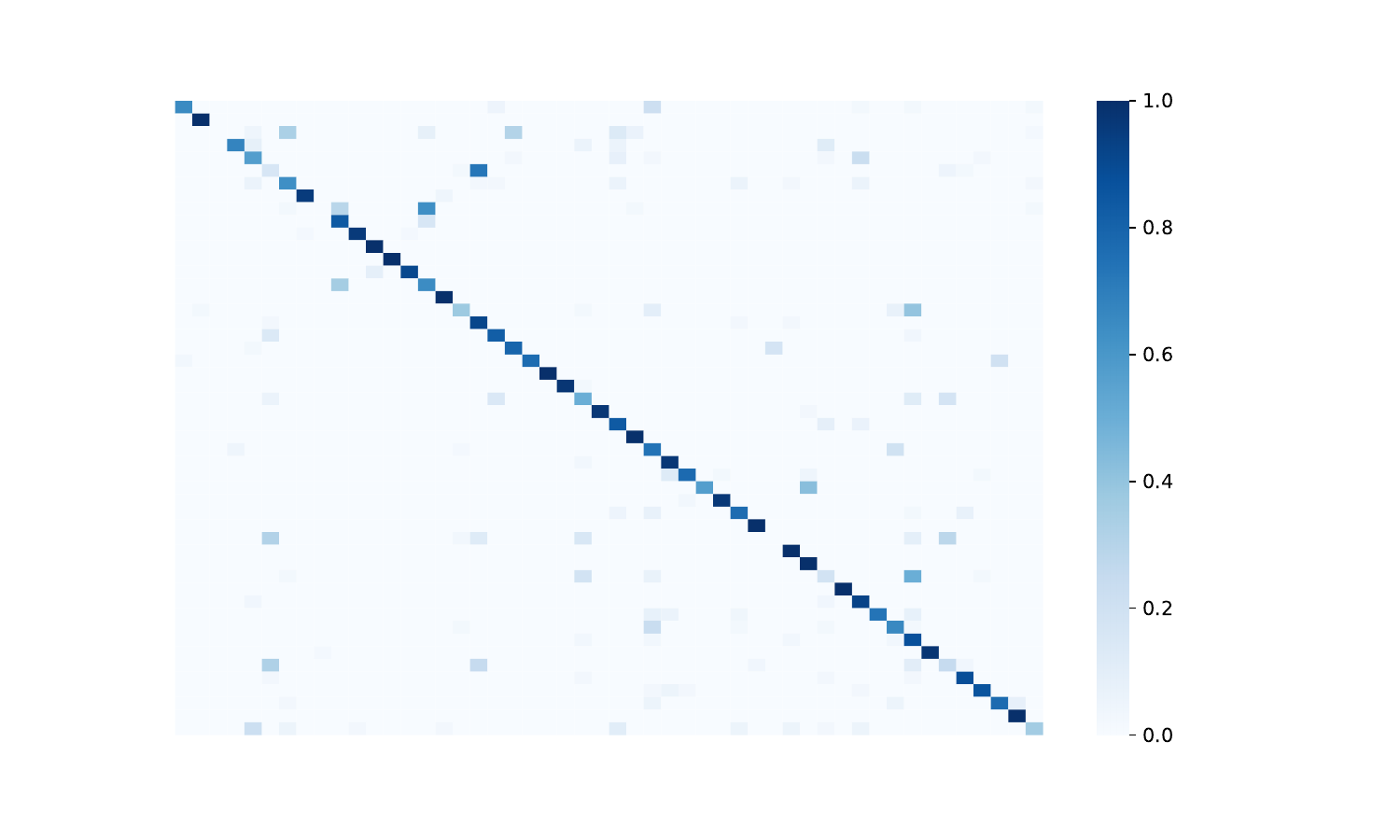}
         \caption{cocoop ucf}
         \label{fig:five over x}
     \end{subfigure}
     \begin{subfigure}[b]{0.3\textwidth}
         \centering
         \includegraphics[width=\textwidth]{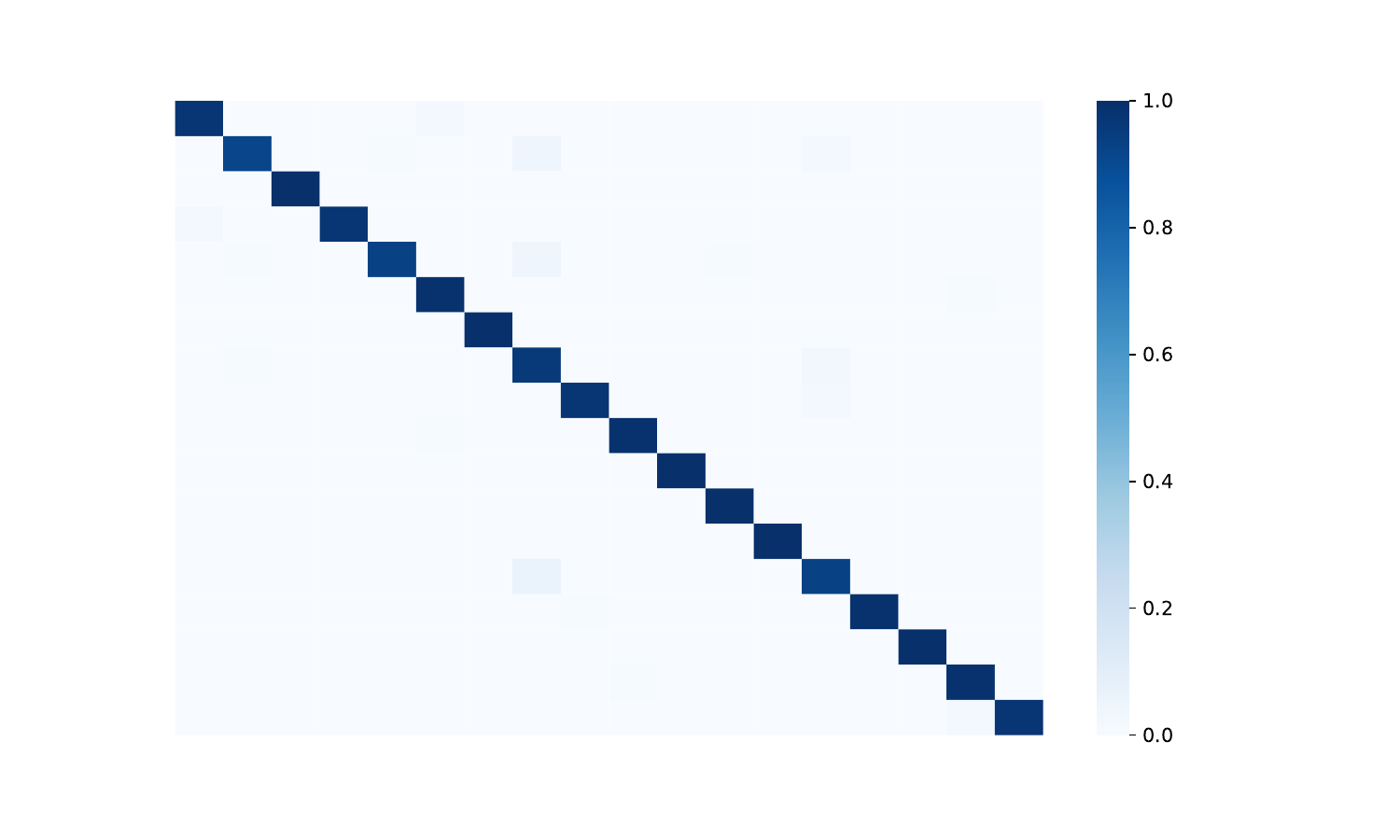}
         \caption{cocoop pets}
         \label{fig:five over x}
     \end{subfigure}
     \caption{The confusion matrix of CoCoOp method on 9 datasets. axis-x stands for each class, while axis-y stands for the class predicted by the model.}
    \label{cocoop}
\end{figure*}

\begin{figure*}[!h]
     \centering
     \begin{subfigure}[b]{0.3\textwidth}
         \centering
         \includegraphics[width=\textwidth]{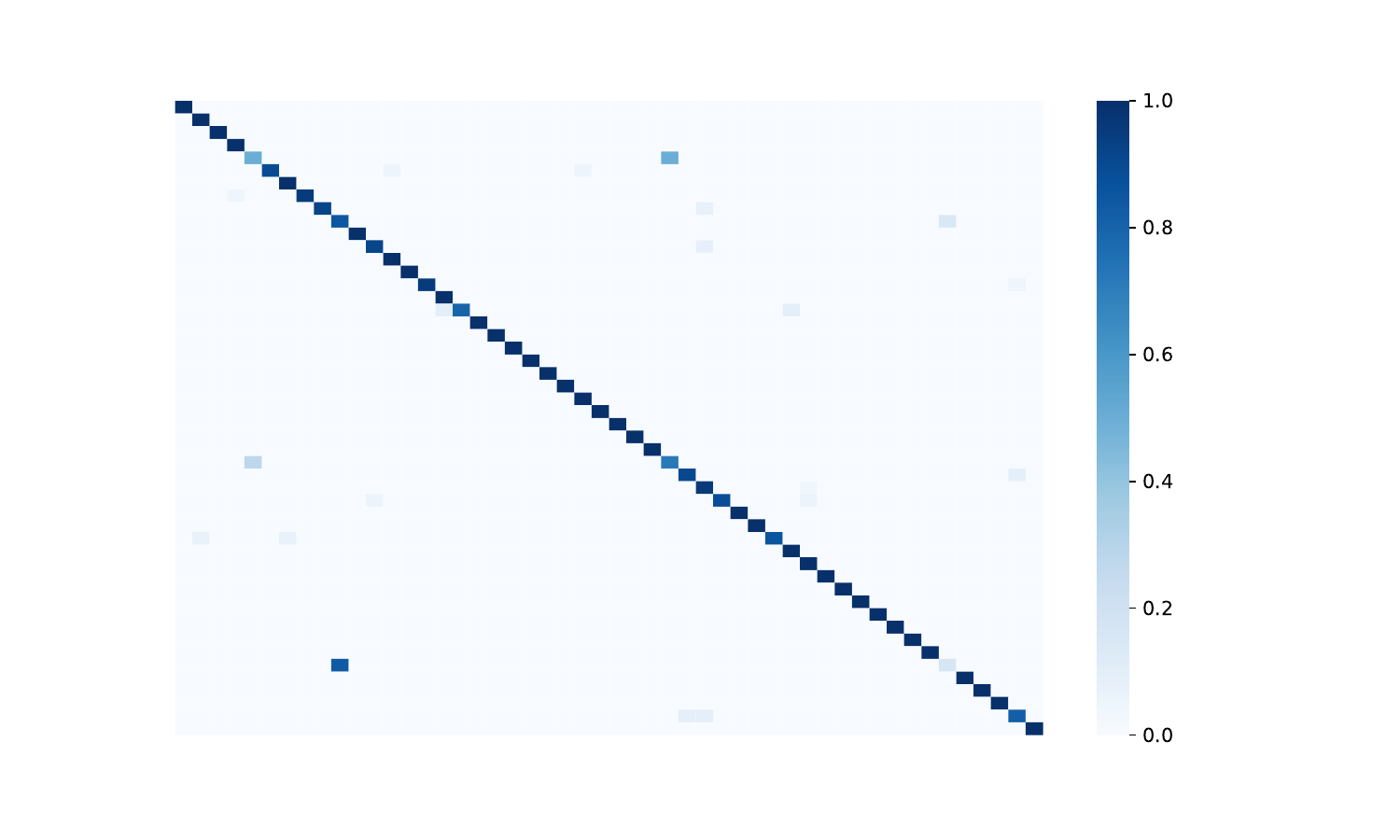}
         \caption{ours caltech101}
         \label{fig:y equals x}
     \end{subfigure}
     \begin{subfigure}[b]{0.3\textwidth}
         \centering
         \includegraphics[width=\textwidth]{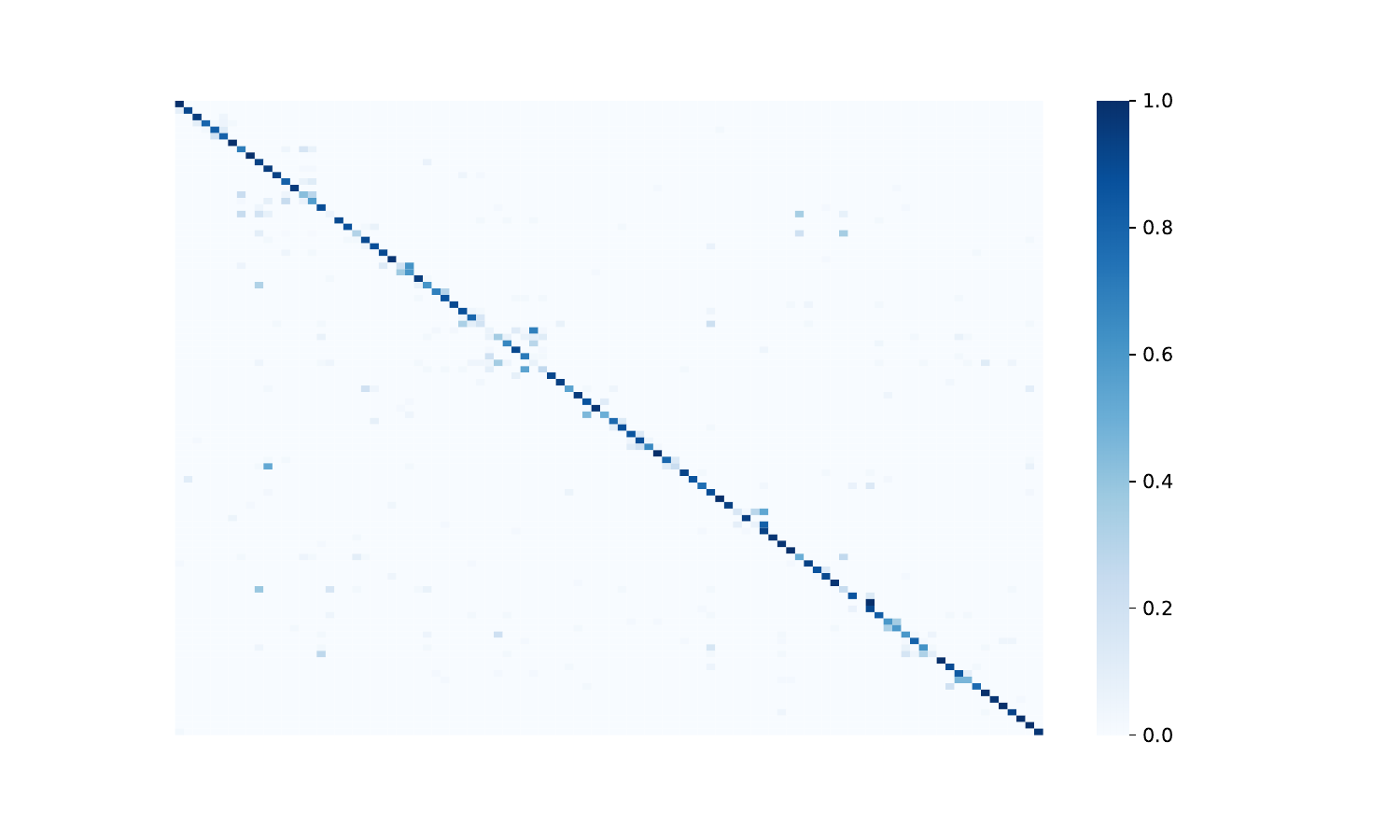}
         \caption{ours cars}
         \label{fig:three sin x}
     \end{subfigure}
     \begin{subfigure}[b]{0.3\textwidth}
         \centering
         \includegraphics[width=\textwidth]{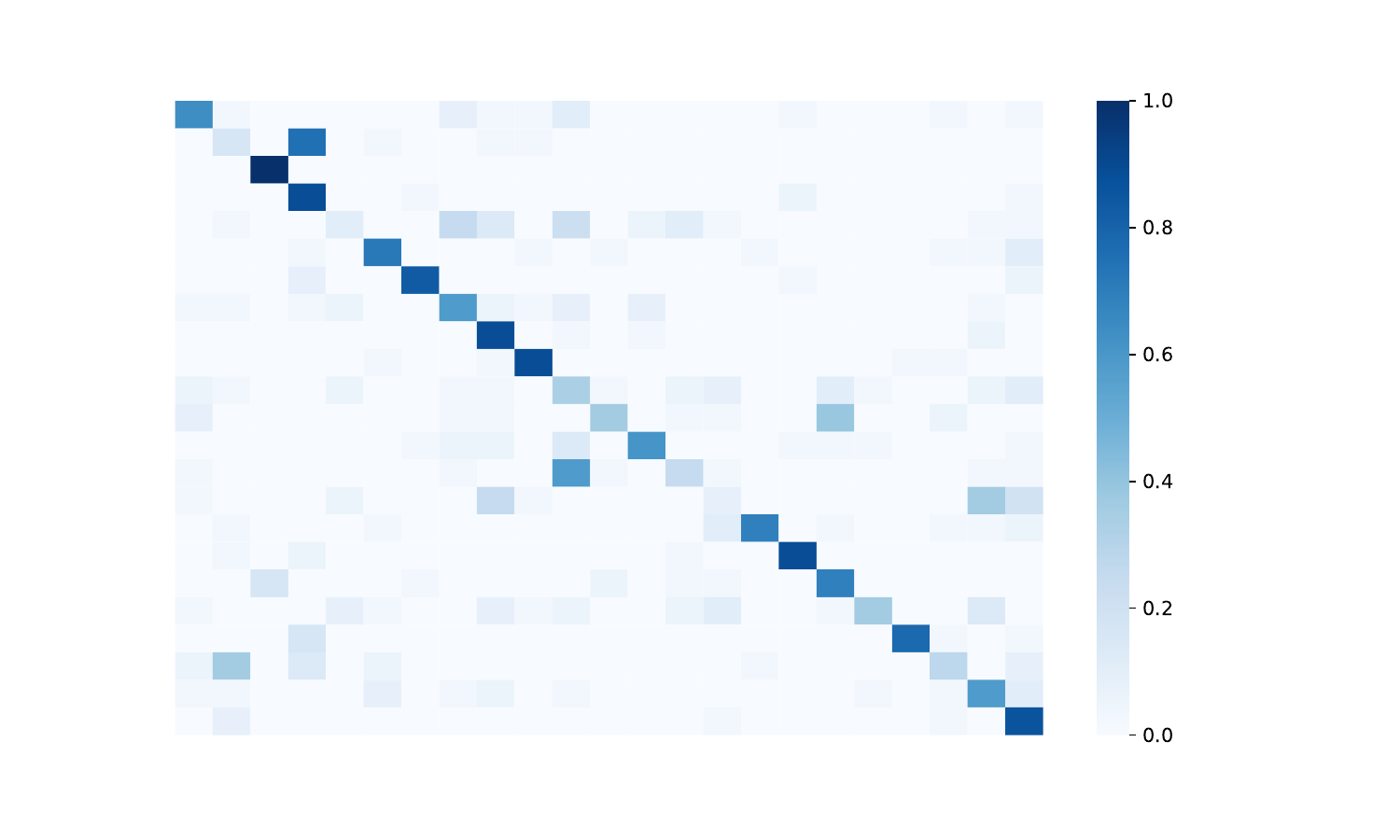}
         \caption{ours dtd}
         \label{fig:five over x}
     \end{subfigure}
     \begin{subfigure}[b]{0.3\textwidth}
         \centering
         \includegraphics[width=\textwidth]{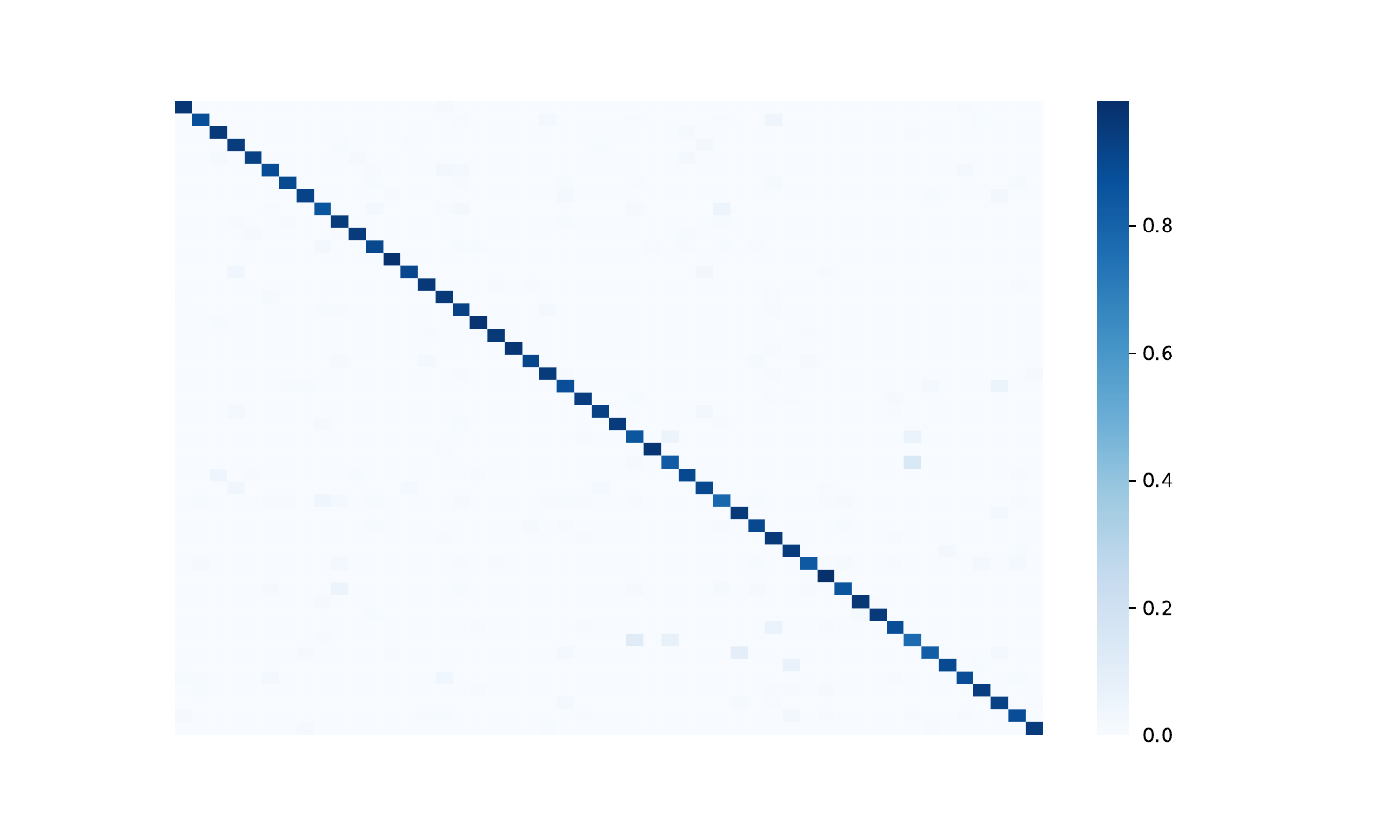}
         \caption{ours food}
         \label{fig:five over x}
     \end{subfigure}
     \begin{subfigure}[b]{0.3\textwidth}
         \centering
         \includegraphics[width=\textwidth]{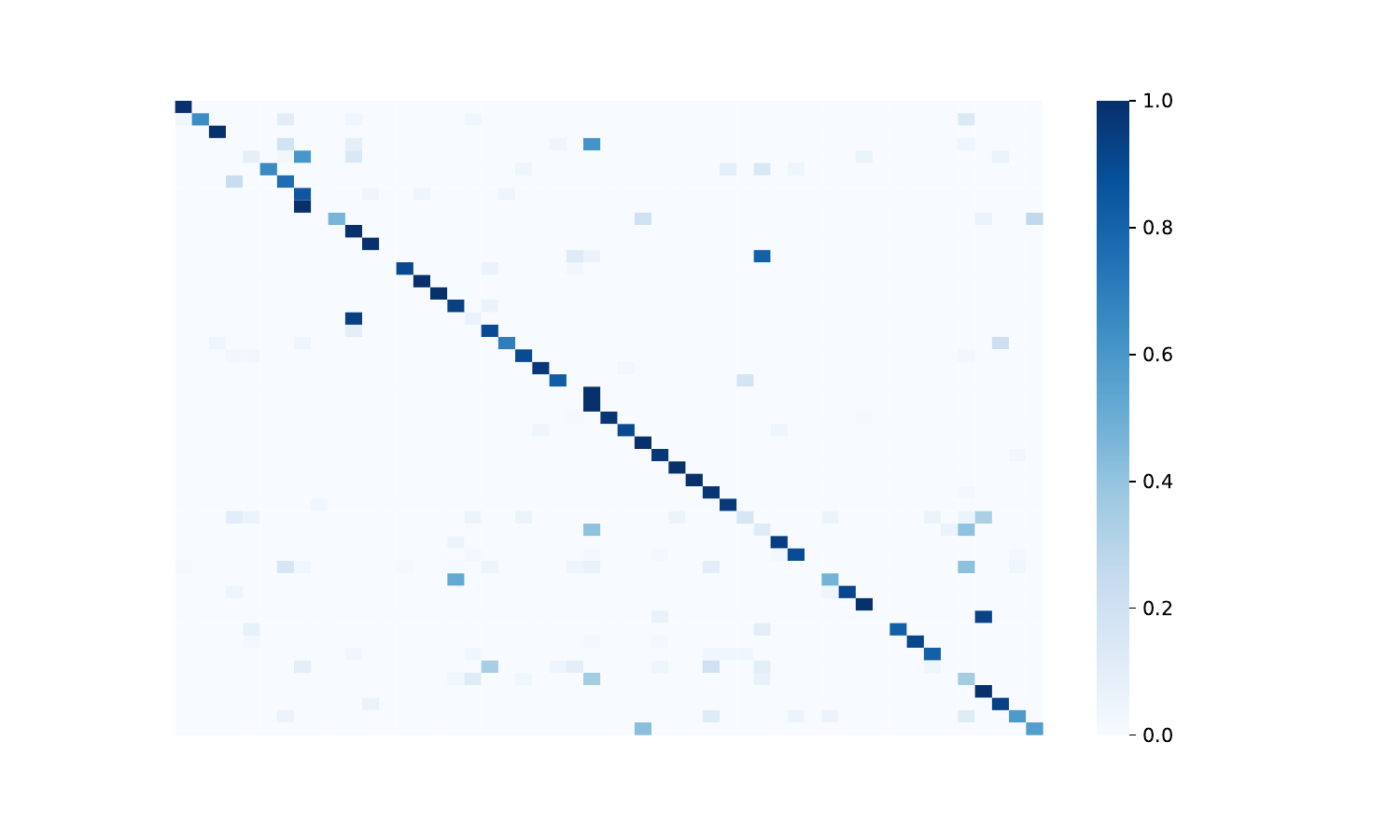}
         \caption{ours flowers}
         \label{fig:five over x}
     \end{subfigure}
     \begin{subfigure}[b]{0.3\textwidth}
         \centering
         \includegraphics[width=\textwidth]{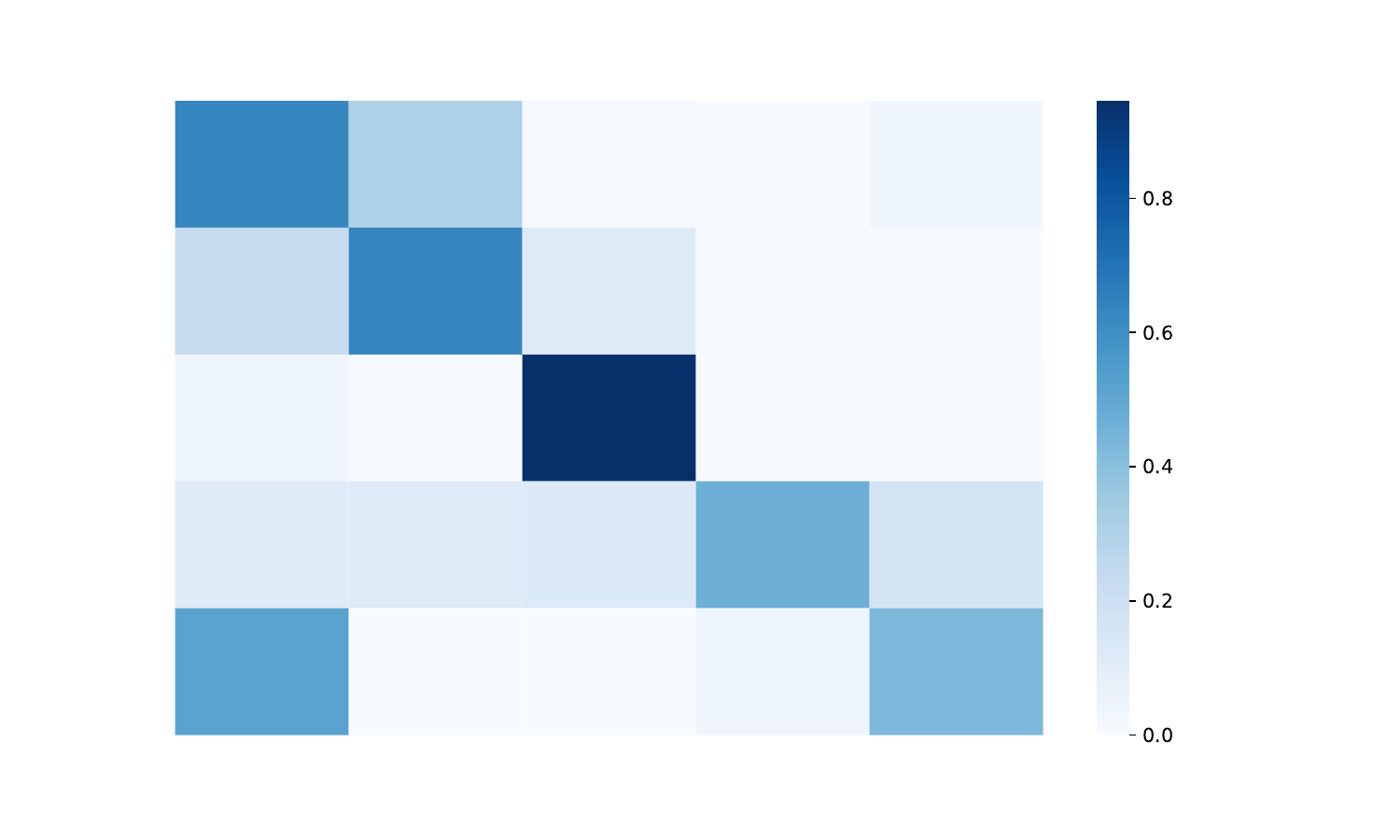}
         \caption{ours eurosat}
         \label{fig:five over x}
     \end{subfigure}
     \begin{subfigure}[b]{0.3\textwidth}
         \centering
         \includegraphics[width=\textwidth]{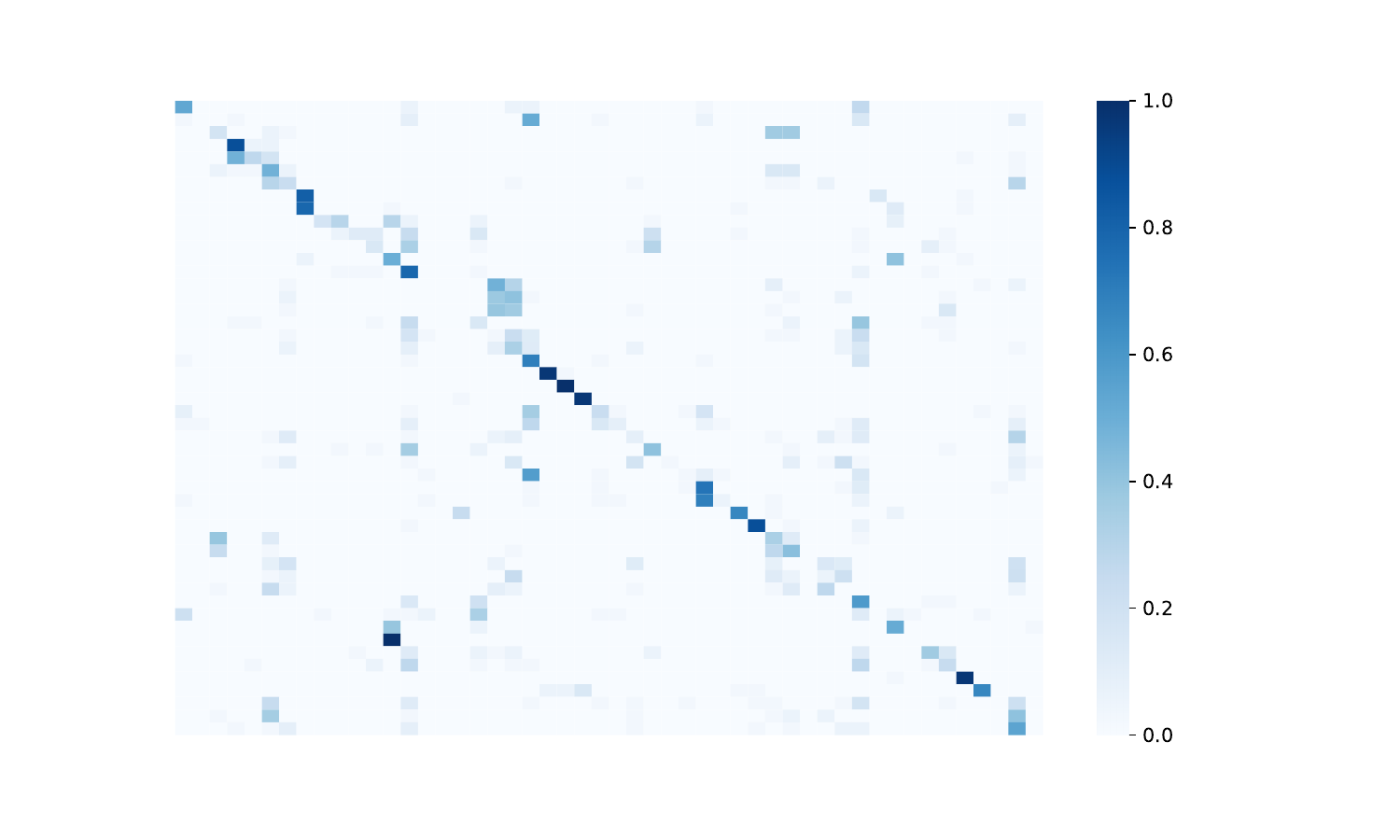}
         \caption{ours aircraft}
         \label{fig:five over x}
     \end{subfigure}
     \begin{subfigure}[b]{0.3\textwidth}
         \centering
         \includegraphics[width=\textwidth]{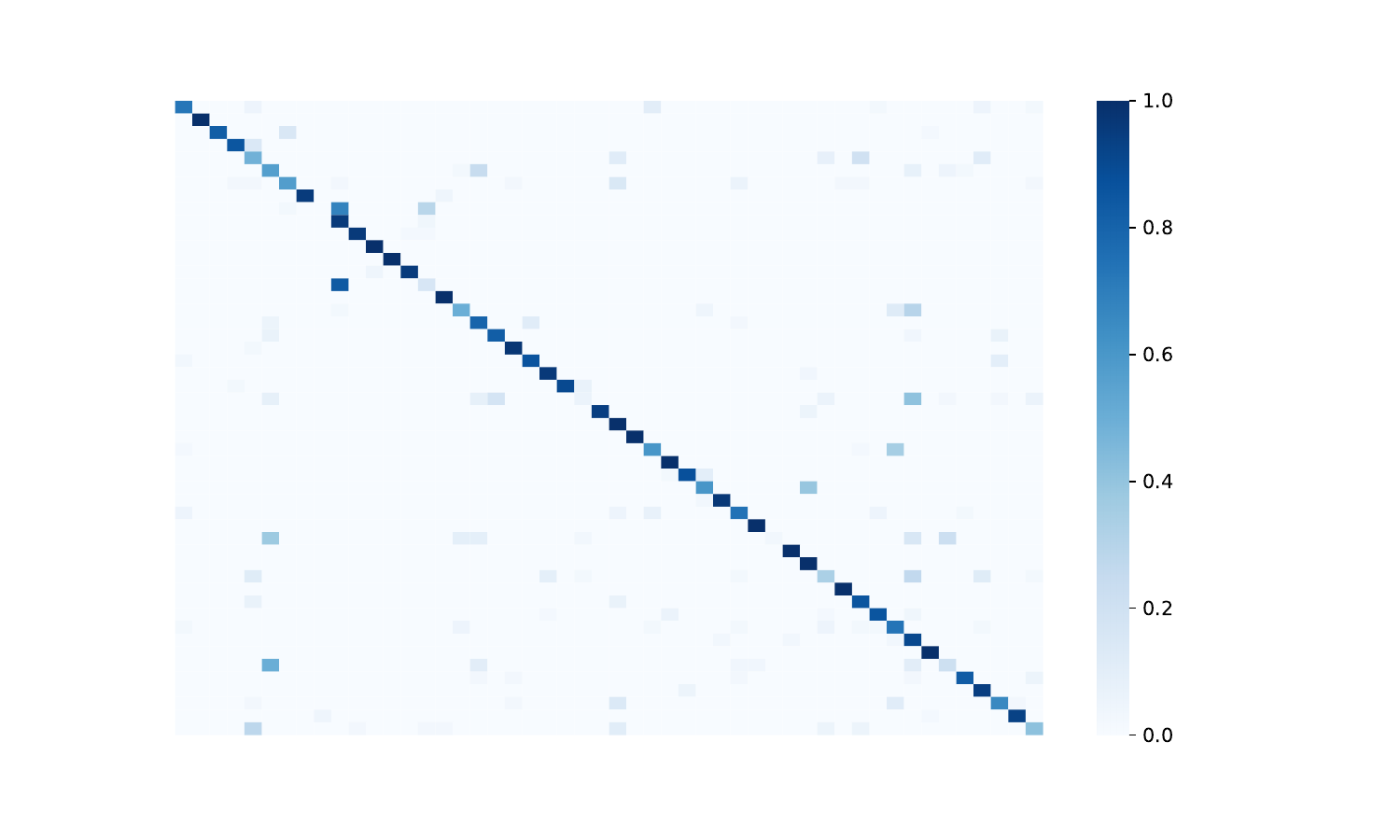}
         \caption{ours ucf}
         \label{fig:five over x}
     \end{subfigure}
     \begin{subfigure}[b]{0.3\textwidth}
         \centering
         \includegraphics[width=\textwidth]{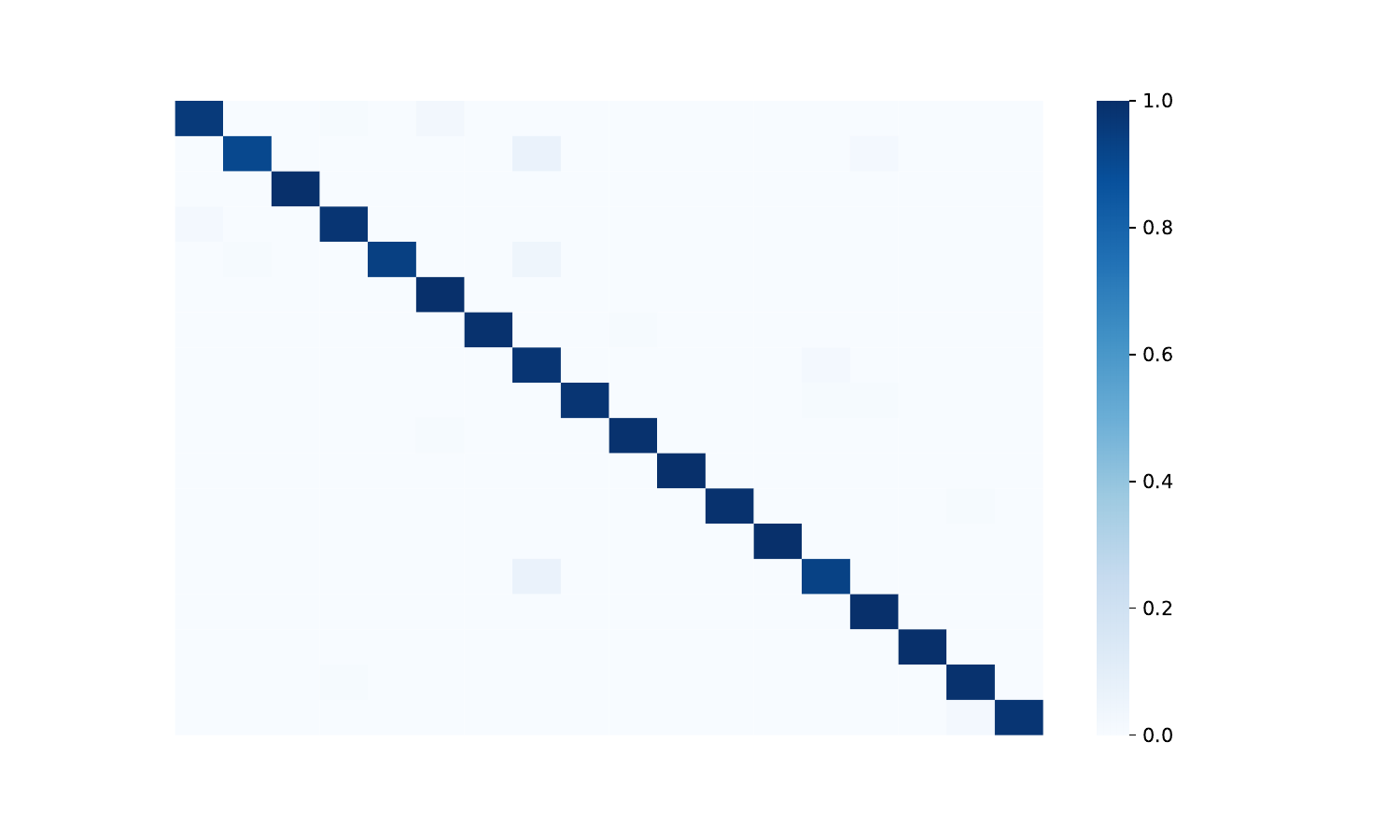}
         \caption{ours pets}
         \label{fig:five over x}
     \end{subfigure}
      \caption{The confusion matrix of our chain of thought method on 9 datasets. axis-x stands for each class, while axis-y stands for the class predicted by the model.} 
    \label{cot}
\end{figure*}

\subsection{Confusion Matrix Visualization}
We provide the confusion matrix visualization of each dataset in Figures \ref{cocoop} and \ref{cot}, where axis-x represents each class and axis-y represents the model's prediction. A dark color on (X, Y) suggests that the model is likely to forecast class X as class Y, whereas a light color on (X, Y) shows that the model is capable of distinguishing between classes X and Y. A good model should have a dark hue on the diagonal and a light color everywhere else. Our method makes the output more focused on the diagonal. demonstrating that our method can help the model discriminate between classes
The confusion matrix may also indicate which classes the model is likely to mix up, which may give insight into how to improve the model's performance.

\section{Additional implementation Details}
\subsection{Implementation details}
In Table \ref{implementation}, we provide further hyper-parameter details. We use a learning rate of 0.002 (identical with CoCoOp) and train the model for 10 epochs. The prompt length is set to 4 for each prompt, initialized with the context "a photo of a". In each Meta-Net, the linear layer is of shape (dim, dim // 16), where dim is the output dim from the image encoder. In the dynamic chain controller, the first linear layer is also of shape (dim, dim//16), and the second linear layer is (dim // 16, n), where n is the chain length.
\subsection{Training and testing details}
\paragraph{Image Classification}
For base to new classification and cross dataset-transfer, we choose the datasets consistent with CoCoOp, covering a wide range of recognization tasks. ImageNet\cite{Imagenetfeifei} and Caltech101\cite{caltech101} are for generic object classification, DTD\cite{dtd} is for texture classification, EuroSAT\cite{eurosat} is for satellite recognition,  StanfordCars\cite{stanfordcars}, FGVCAircraft\cite{aircraft}, Flowers102\cite{flowers102}, and Food101\cite{food} are for fine-grained classification, SUN397\cite{sun397} is for scene recognition, and UCF101\cite{ucf101} is for action recognition. For domain generalization, we also choose the same datasets with CoCoOp, including ImageNetV2\cite{imagenetv2}, which contains new test data for the ImageNet\cite{Imagenetfeifei} benchmark, containing three test sets with 10,000 new images each; ImageNet-A\cite{imageneta} that consists of real-world, unmodified, and naturally occurring examples misclassified by ResNet models; ImageNet-R\cite{imagenetr} that contains art, cartoons, deviantart, graffiti, embroidery, graphics, origami, paintings, patterns, plastic objects, plush objects, sculptures, sketches, tattoos, toys, and video game renditions of ImageNet classes; and finally ImageNet-sketch\cite{imagenetsketch} that contains only "black and white" color scheme.

\paragraph{Image Text Retrieval}
We use tow datasets: MSCOCO\cite{mscoco}, which is a large-scale object detection, segmentation, key-point detection, and captioning dataset with a total of 123K images, and Flickr30k\cite{Retrieval}, which contains 31K images in total, each provided with 5 sentences annotated by human annotators. We sample 0.5\%, 1.0\%, and 1.5\% from the training dataset to create the few-shot setting, and we fix the seed to 1 while spliting. We use the first caption of the image as its prompt.
\paragraph{Visual Question Answering}
Visual Question Answering (VQA) v2.0\cite{VQA} is a challenging dataset containing open-ended questions about images. Given an image and a natural language query, the model is asked to reason about vision, language, and common sense in order to come up with the correct answer. We use the official split of the VQA v2 with 83k images and 444k questions for training and 41k images and 214k questions for validation. The model picks the corresponding answer from a set of 3,129 answers and treats it as classification for each question. Due to GPU limitation, we sample 0.25\%, 0.5\%, and 0.75\% of the training set for the few-shot setting (instead of 0.5\%, 1.0\%, 1.5\% as in Image-Text Retrieval), fixing seed to 1 for sampling.
We use the concatenation of question-answer pairs as the instance-specific prompts.

\end{document}